\documentclass{article}

\usepackage{arxiv}

\usepackage[utf8]{inputenc} 
\usepackage[T1]{fontenc}    
\usepackage{hyperref}       
\usepackage{url}            
\usepackage{booktabs}       
\usepackage{amsfonts}       
\usepackage{nicefrac}       
\usepackage{microtype}      
\usepackage{lipsum}
\usepackage{graphicx}
\usepackage[numbers]{natbib}
\usepackage{doi}

\usepackage[ruled,vlined]{algorithm2e}

\usepackage{amsmath,amsfonts,bm}

\def\Figref#1{Figure~\ref{#1}}

\def\eqref#1{equation~\ref{#1}}
\def\Eqref#1{Equation~\ref{#1}}

\def\1{\bm{1}}

\DeclareMathAlphabet{\mathsfit}{\encodingdefault}{\sfdefault}{m}{sl}
\SetMathAlphabet{\mathsfit}{bold}{\encodingdefault}{\sfdefault}{bx}{n}

\usepackage{wrapfig}
\usepackage{amsmath}
\usepackage{subcaption} 
\usepackage{xcolor}         
\usepackage[table]{xcolor}

\usepackage[english]{babel}

\usepackage{xspace}
\usepackage[varqu,varl,var0,scaled=0.97]{inconsolata}
\usepackage{caption}
\usepackage{overpic}
\usepackage{wrapfig}
\usepackage{tcolorbox}
\usepackage{enumitem}
\usepackage{threeparttable}
\usepackage{amssymb,amsmath,amsthm}
\usepackage{listings}
\usepackage{lipsum}
\usepackage{multicol}
\usepackage{multirow}
\usepackage{hyperref}
\usepackage{color}
\definecolor{dkgreen}{rgb}{0,0.6,0}
\definecolor{gray}{rgb}{0.5,0.5,0.5}
\definecolor{mauve}{rgb}{0.58,0,0.82}

\newcommand{\eg}{\emph{e.g.},\xspace}

\definecolor{ao(english)}{rgb}{0.0, 0.5, 0.0}

\usepackage[capitalize,noabbrev]{cleveref}

\newcommand{\hi}[1]{\noindent{\textcolor{violet}{\textbf{#1}}}}

\theoremstyle{plain}

\newcommand{\name}{\textsc{GRAIL}\xspace}

\title{\name: Post-hoc Compensation by Linear Reconstruction for Compressed Networks}

\author{
Wenwu Tang\textsuperscript{$\diamond$},
Dong Wang\textsuperscript{$\diamond$}, 
Lothar Thiele\textsuperscript{$\star$}, 
Olga Saukh\textsuperscript{$\diamond$,§} \\
\textsuperscript{$\diamond$}Graz University of Technology, 
\textsuperscript{§}Complexity Science Hub, Austria \\
\textsuperscript{$\star$}ETH Zurich, Switzerland \\
\texttt{\{wenwu.tang, dong.wang, saukh\}@tugraz.at}, 
\texttt{thiele@tik.ee.ethz.ch} \\
}

\begin{document}
\maketitle

\begin{abstract}
Structured deep model compression methods are hardware-friendly and substantially reduce memory and inference costs. However, under aggressive compression, the resulting accuracy degradation often necessitates post-compression finetuning, which can be impractical due to missing labeled data, or high training cost.
We propose \emph{post-hoc blockwise compensation}, called \name, a simple zero-finetuning step applied after model compression that restores each block’s input–output behavior using a small calibration set. The method summarizes hidden activations via a Gram matrix and applies ridge regression to linearly reconstruct the original hidden representation from the reduced one. The resulting reconstruction map is absorbed into the downstream projection weights, while the upstream layer is compressed. The approach is selector-agnostic (Magnitude, Wanda, Gram-based selection, or folding), data-aware (requiring only a few forward passes without gradients or labels), and recovers classic pruning/folding when the Gram matrix is near identity, indicating weak inter-channel correlations. Across ResNets, ViTs, and decoder-only LLMs, \name consistently improves accuracy or perplexity over data-free and data-aware pruning/folding baselines in practical compression regimes, with manageable overhead and no backpropagation. Our code is available at: \href{https://github.com/TWWinde/GRAIL_Compensation}{https://github.com/TWWinde/GRAIL}
\end{abstract}

\section{Introduction}

Deployed models increasingly face tight latency and memory budgets, and achieving high-quality compression is essential for meeting these constraints without sacrificing accuracy. Yet in many deployment settings, retraining or even light finetuning after compression is infeasible: data may be proprietary, labels unavailable, or training cycles too costly for the post-training phase~\citep{choudhary2020effcientNNsurvey,frantar2023gptq,wang2025forget}. This motivates training-free, weight-space compression methods that directly narrow hidden width. Two structured families dominate: pruning, which selects a subset of channels or heads~\citep{yang2025wandapruninglargelanguage,frantar2023sparsegptmassivelanguagemodels}, and folding, which clusters and merges them into a smaller basis~\citep{wang2025forget,chen2024ifm}. Both pruning and folding modify the hidden representation seen by the next layer, yet typically do so without accounting for how dropped units contribute under the actual data distribution. In practice, reducing layer width changes the downstream representation statistics and can degrade accuracy or perplexity.

We address this gap with a simple \emph{post-hoc blockwise compression compensation} step, called \name, that restores the block’s input-output behavior without labels or training. In this paper, we refer to the layer whose outputs are compressed as the \textit{producer}, and the immediately downstream layer that consumes these activations as the \textit{consumer}. After a pruning or folding decision has been made (by any criterion), we briefly run the uncompressed model in evaluation mode on a small, unlabeled calibration set and record the activations at the consumer input (post-activation inputs for dense layers, concatenated per-head features just before the attention output projection). From these activations we compute only second-order statistics (Gram matrices). Using this summary, we fit a linear map
that predicts the original hidden from the reduced hidden. Finally, we absorb this map into the consumer projection weights, while the producer is narrowed by selection (pruning) or per-cluster averaging (folding). \name does not attempt to push a linear map through a nonlinearity. Instead, the reconstruction is explicitly learned after the non-linear activation. The result is a one-shot, data-aware correction that requires no gradients, no labels, and adds only a small linear solve per block.

Our framework is post hoc and modular: it is compatible with various model compression methods such as magnitude pruning, Wanda~\citep{yang2025wandapruninglargelanguage}, SlimGPT~\citep{ling2024slimgpt}, FLAP~\citep{yongqian2024flap}, or clustering-based folding~\citep{wang2025forget}. It uses only a small amount of calibration data, introduces low runtime overhead and zero additional parameters. 
Across ResNets~\citep{he2015resnets}, ViTs~\citep{dosovitskiy2021vits}, and decoder-only LLMs, \name consistently improves over pruning-only and folding-only baselines, narrowing the gap to the finetuned model without any finetuning. Our contributions are:
\begin{itemize}
\item \textbf{Training-free post-hoc blockwise compensation.} 
We propose a data-aware, one-shot linear weight compensation applied \emph{after} structured model compression (pruning or folding) that restores each block’s input–output behavior using a small calibration dataset to compute second-order activation statistics---no labels, no additional parameters, without calculating gradients and finetuning.
\item \textbf{Practical recipe and broad evidence.} 
Large-scale experiments on ResNets~\citep{he2015resnets}, ViTs~\citep{dosovitskiy2021vits}, and decoder-only LLMs across more than 700 checkpoints on CIFAR-10~\citep{cifar10}, ImageNet-1K~\citep{deng2009imagenet}, and C4~\citep{raffel2020exploring} / WikiText-2~\citep{merity2016pointer} / PTB~\citep{marcus-etal-1993-building} show consistent gains over structured model compression baselines. Our method also improves upon strong structured compression baselines that include their own recovery mechanisms, such as FLAP~\citep{yongqian2024flap} with built-in bias correction, SlimGPT~\citep{ling2024slimgpt} curvature-based reconstruction and Wanda++~\citep{yang2025wanda} with local optimization.
\item \textbf{Low data requirement, memory and compute overhead.} 
The method requires only a tiny, unlabeled calibration data from training set (Hundreds of images or tens of thousands of tokens), no gradients, and an efficient Gram accumulation, requiring  manageable memory and time even for CLIP~\citep{radford2021learningtransferablevisualmodels} and LLaMA-2-7B~\citep{touvron2023llama2openfoundation} models.
\end{itemize}

\begin{figure}[t]
    \centering
    \includegraphics[width=0.95\linewidth]{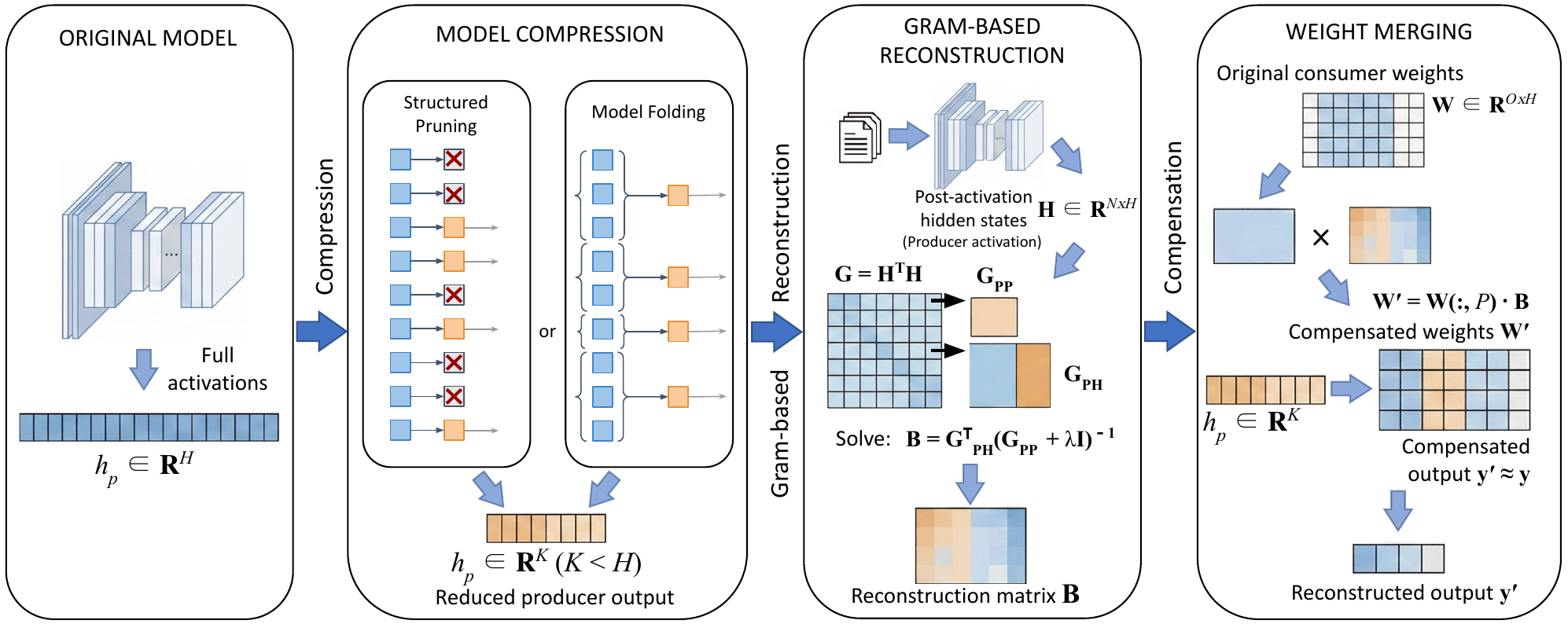}
    \caption{\textbf{\name: GRAm-Integrated Linear compression workflow.} 
    After a structured compression decision (pruning / folding) narrows the producer’s hidden width, we run a small, unlabeled calibration set through the upstream blocks and collect activations at consumer input. From these activations, \name forms the Gram matrix of second-order statistics and solves a ridge regression that reconstructs the original hidden representation from the reduced one. The resulting linear compensation is merged into the consumer projection weights, while the producer is narrowed by selection or clustering. This one-shot, training-free step restores each block’s input–output behavior and applies uniformly to pruning and folding across CNNs, ViTs, and LLMs.}
    \label{fig:figureone}
\end{figure}

\section{Related Work}
\label{sec:related}

\paragraph{Closest compensation-related works.}
Compensation in structured width reduction is typically limited or implicit.  
Pruning methods often retain surviving channels without correcting the downstream projection, relying on fine-tuning or localized updates to regain accuracy \citep{he2017channelpruningacceleratingdeep,lee2019snipsingleshotnetworkpruning,michel2019sixteenheadsreallybetter}.  
Folding and clustering approaches provide an algebraic update of the consumer weights \citep{wang2025forget,chen2024ifm,saukh2025cutlessfoldmore}, but these updates are data-free and assume that removed units are well represented by simple centroids, which may diverge from the true activation geometry.  
Second-order methods (OBS/OBD) introduce principled loss-aware corrections \citep{Hassibi1993OBS,LeCun1990OBD}, yet they primarily target unstructured parameter removal and do not enforce architectural constraints such as multi-head attention reshape rules. SparseGPT \citep{frantar2023sparsegptmassivelanguagemodels} decomposes post-training compression into single-layer subproblems, where the objective is to minimize the $\ell_2$ error between the outputs of an uncompressed layer and its compressed counterpart. Reconstruction is performed within the pruned layer itself, directly updating its weights under a fixed pruning mask. But it was originally proposed for unstructured pruning, and the weight selection and weight update are tightly coupled. While SparseGPT demonstrated the feasibility of efficient post-training unstructured pruning, ZipLM \citep{frantar2023sparsegptmassivelanguagemodels} extends these principles to the structured setting, incorporating inference-aware constraints to ensure realizable runtime speedups on standard hardware without the need for specialized sparse kernels. A closely related work is channel pruning with output reconstruction \citep{he2017channelpruningacceleratingdeep}, which iteratively removes input channels of a single layer and solves a least-squares problem to preserve that layer’s output. However, it operates locally on individual projections, requires repeated pruning–reconstruction steps during or after training, and does not treat coupled producer–consumer pairs such as the $(W_{\mathrm{fc}},W_{\mathrm{proj}})$ structure in Transformers. 

Recent LLM-focused pruning methods such as SlimGPT~\citep{ling2024slimgpt} and Wanda++~\citep{yang2025wanda} incorporate local reconstruction or gradient-based corrections, and FLAP~\citep{yongqian2024flap} adds head-wise bias compensation, but these mechanisms are tied to specific pruning heuristics and are not general-purpose compensation methods. Moreover, these approaches are designed for decoder-only Transformers and do not naturally transfer to CNNs, ViTs, or folding-based width reduction. 
Recent work has also begun to explore post-pruning recovery for LLMs: \eg \citet{shen2025numericalpruning} introduce a Newton-based structured pruning method with a built-in compensation step, while \citet{feng2025restoringprunedlargelanguage} propose re-injecting lost attention components to restore performance in pruned LLMs. 
Our work differs in that \name provides a unified, linear reconstruction which is selector-agnostic, training-free, and applies across both structured pruning and folding, as well as to vision and language models.

\paragraph{Activation-matching compensation.}
The most direct analogues are PTQ reconstruction methods, which match layer outputs on calibration data to compensate quantization errors~\citep{nagel2020downadaptiveroundingposttraining,li2021brecq,frantar2023gptq, lin2024awqactivationawareweightquantization}.  
These approaches optimize rounding or scaling and can incorporate second-order guidance, but they operate at fixed width and therefore do not handle dimension-changing reducers from pruning or folding, nor head-structured transformations in LLMs.  
Distillation-based techniques~\citep{hinton2015distillingknowledgeneuralnetwork} also provide activation alignment but require full student training.  
Overall, prior work lacks a unified, training-free, activation-aware compensation specifically designed for structured width reduction.

\section{\name: GRAm-Integrated Linear compensation}
\label{sec:method}

Most classical structured model compression methods reduce channels under the implicit assumption that removing certain hidden units has a negligible impact on the remaining ones. This weight–space heuristic often distorts the consumer’s input geometry, especially in pre-LN Transformers where hidden activations co-activate in structured ways~\citep{sun2024simpleeffectivepruningapproach,dettmers2022llmint88bitmatrixmultiplication}. As illustrated in \Figref{fig:figureone}, \name replaces this assumption with a data-aware correction step: using a small, unlabeled calibration set, we accumulate the consumer-input Gram matrix and solve a ridge regression that reconstructs the original hidden representation from its reduced counterpart. The resulting linear map is then merged into the consumer weights,
yielding a selector-agnostic, zero-finetuning procedure that restores the block’s input–output behavior. Below we detail this procedure and show how it applies uniformly across CNN, ViT, and LLM blocks.

\subsection{Dense, convolutional and FFN blocks}

We consider a block consisting of a producer layer \(i{-}1\) followed by a consumer layer \(i\). Let \(h\) denote the output activation from the producer, which serves as the input to the consumer. \name learns a cross-layer compensation from producer activations, then folds it into the
consumer weights.

\paragraph{Dense block compensation.}
Consider two consecutive fully connected layers \(i{-}1\) and \(i\):
\[
h = \phi(\mathbf{W}_{\mathrm{i-1}} z + b_{\mathrm{i-1}}) \in \mathbb{R}^{H}, \qquad
y = \mathbf{W}_{\mathrm{i}} h + b_{\mathrm{i}} \in \mathbb{R}^{O},
\]
where \(\mathbf{W}_{\mathrm{i-1}} \in \mathbb{R}^{H \times C}\),
\(\mathbf{W}_{\mathrm{i}} \in \mathbb{R}^{O \times H}\) and \( \phi \) denotes activation function.
We compress the hidden dimension from \(H\) to \(K \ll H\) by selecting a subset \( P \subset \{1,\dots,H\} \) of
size \(|P| = K\) or using model folding. The selection step is method-agnostic: \(P\) may be obtained
using activation norms, weight magnitudes, Wanda~\citep{yang2025wandapruninglargelanguage}, or Gram-based selection~\citep{Sarvani_2022}.
In folding methods, channels are grouped into clusters and each cluster is replaced by its centroid~\citep{wang2025forget}. The reduced basis \(P\) consists of these centroids, with all channels collapsed onto their cluster means.

Many model compression methods can be interpreted as applying a linear mapping
matrix that reduces the hidden width~\citep{saukh2025cutlessfoldmore}. Let
\(\mathbf{H} \in \mathbb{R}^{N \times H}\) denote the original hidden activations where each row corresponds to an activation $h$ for one of $N$ data samples. Width reduction for target dimension \(K < H\) can be written universally as
\[
\mathbf{H}_{\mathrm{red}} = \mathbf{H} \mathbf{M},
\qquad
\mathbf{M} \in \mathbb{R}^{H \times K}.
\]

In \emph{structured pruning}, \(\mathbf{M}\) is a binary matrix that 
retains only a subset of channels indexed by \(P\):
\[
\mathbf{M}_{\mathrm{prune}} = 
\begin{bmatrix}
\mathbf{e}_{p_1} & \mathbf{e}_{p_2} & \cdots & \mathbf{e}_{p_K}
\end{bmatrix},
\qquad
\mathbf{H}_{\mathrm{pruned}} = \mathbf{H}\,\mathbf{M}_{\mathrm{prune}},
\]
where \(\mathbf{e}_{p_i}\) are standard basis vectors. 
In contrast, \emph{model folding} groups channels into $K$ clusters $C_k$, $ 1 \leq k \leq K$ of activations and replaces each cluster by its
centroid. The folding map
\[
\mathbf{M}_{\mathrm{fold}}(h,k)=
\begin{cases}
1/|C_k|, & h \in C_k,\\[2pt]
0, & \text{otherwise},
\end{cases}
\qquad
\mathbf{H}_{\mathrm{folded}} = \mathbf{H}\,\mathbf{M}_{\mathrm{fold}},
\]
where $h$ denotes the index of an activation averages all channels in \(C_k\) into a single direction.  Thus \(\mathbf{M}_{\mathrm{fold}}\) is a
low-rank projection that mixes channels rather than discarding them.

\name learns a data-aware cross-channel
reconstruction mapping \(\mathbf{B}\) from post-activation hidden statistics, and merges this reconstruction into the consumer weight \(\mathbf{W}_{\mathrm{i}}\), improving
compressed models without finetuning. Our goal is to approximate the original hidden representation \(h\) using only the retained subset \(h_P\). We seek a linear reconstruction map \(\mathbf{B} \in \mathbb{R}^{H \times K}\) such that
\[
h \;\approx\; \mathbf{B} h_P
\]
on a small unlabeled calibration set. By collecting activations into matrices
\(\mathbf{H}\) and \(\mathbf{H}_P\) across \(N\) samples, we solve the ridge regression problem:
\[
\min_{\mathbf{B}}
\; \big\| \mathbf{H} - \mathbf{H}_P \mathbf{B}^\top \big\|_F^2
+ \lambda \|\mathbf{B}\|_F^2,
\]
where \(\lambda > 0\) is a ridge regularization coefficient that stabilizes the
solution when \(\mathbf{H}_P^\top \mathbf{H}_P\) is ill-conditioned
or near-singular. The normal equations give the closed-form solution:
\[
\mathbf{B}
= \mathbf{H}^\top \mathbf{H}_P
  \left(\mathbf{H}_P^\top \mathbf{H}_P + \lambda \mathbf{I}\right)^{-1}.
\]

For pruning, the reducer is a selection matrix, and therefore \(\mathbf{G}_{PP} = \mathbf{G}[P,P]\) is the correct Gram submatrix. For folding, however, the reducer is a merge map \(\mathbf{M}_{\mathrm{fold}} \in \mathbb{R}^{H \times K}\), so the reduced activations satisfy \(\mathbf{H}_{\mathrm{red}} = \mathbf{H} \mathbf{M}_{\mathrm{fold}}\). Consequently, the Gram block entering the ridge system becomes
\[
\mathbf{G}_{PP}^{\mathrm{fold}}
= \mathbf{H}_{\mathrm{red}}^\top \mathbf{H}_{\mathrm{red}}
= \mathbf{M}_{\mathrm{fold}}^\top\, \mathbf{G}\, \mathbf{M}_{\mathrm{fold}},
\]
which generalizes the pruning case and correctly accounts for channel mixing.
Substituting this expression into the reconstruction formula preserves the unified form \(\mathbf{B} = \mathbf{G}_{PH}^\top (\mathbf{G}_{PP} + \lambda \mathbf{I})^{-1}\).

The reconstruction map \(\mathbf{B}\) is merged into the consumer projection weights:
\[
\mathbf{W}'_{\mathrm{i}} = \mathbf{W}_{\mathrm{i}} \mathbf{B}
\in \mathbb{R}^{O \times K},
\]
while the producer weights are width-reduced by indexing:
\[
\mathbf{W}'_{\mathrm{i-1}} = \mathbf{W}_{\mathrm{i-1}}[P,:], \qquad
b'_{\mathrm{i-1}} = b_{\mathrm{i-1}}[P], \qquad
b'_{\mathrm{i}} = b_{\mathrm{i}}.
\]

At inference, the reduced hidden activations \( h_P = \phi(\mathbf{W}'_{\mathrm{i-1}} z + b'_{\mathrm{i-1}}) \) produce
\[
y' = \mathbf{W}'_{\mathrm{i}} h_P + b'_{\mathrm{i}} \;\approx\; \mathbf{W}_{\mathrm{i}} h + b_{\mathrm{i}}.
\]

An equivalent formulation minimizes
\[
\big\|\mathbf{H}_P \mathbf{W}'_{\mathrm{i}}{}^\top - \mathbf{H} \mathbf{W}_{\mathrm{i}}^\top \big\|_F^2
\;+\; \lambda \|\mathbf{W}'_{\mathrm{i}}\|_F^2,
\]
yielding
\[
\mathbf{W}'_{\mathrm{i}} = \mathbf{W}_{\mathrm{i}} \mathbf{G}_{PH}^\top (\mathbf{G}_{PP} + \lambda \mathbf{I})^{-1},
\]
which is identical to the merged form via the regression solution.

\paragraph{Convolutional block compensation.}
For a convolutional layer with weights
\(\mathbf{W}_{\mathrm{conv}} \in \mathbb{R}^{O \times H \times k_H \times k_W}\),
the compensation map \( \mathbf{B} \in \mathbb{R}^{H \times K} \) learned from the
activation regression of the preceding fully connected case is applied along the
\emph{input channel} dimension of the convolution kernel:
\[
\mathbf{W}^{\mathrm{new}}_{\mathrm{conv}}(o,k,:,:)
=
\sum_{h=1}^{H}
\mathbf{W}_{\mathrm{conv}}(o,h,:,:)\, \mathbf{B}(h,k).
\]

\paragraph{Compensation for ViTs and CLIP.}
For Vision Transformers (ViT) and CLIP models \name operates at the block level, specifically targeting the MLP modules. 
For Transformer MLP/FFN blocks \(\mathbf{W}_{\mathrm{fc}}\) (expansion) and \(\mathbf{W}_{\mathrm{proj}}\) (projection) form a coupled producer–consumer pair. The pruning process removes rows from $\mathbf{W}_{\text{fc}}$ and corresponding columns from $\mathbf{W}_{\text{proj}}$. The compensation is then applied exclusively to the input of the projection layer to reconstruct the output of the original block:
\begin{equation}
    \mathbf{W}'_{\text{proj}} = \mathbf{W}_{\text{proj}} \mathbf{B},
\end{equation}
where $\mathbf{B}$ is the regression matrix derived from the Gram statistics of the hidden activations. This coupled treatment ensures that the interaction between the two layers is preserved, which is critical for maintaining the expressivity of the MLP.

\paragraph{Complexity.}
Accumulating \(\mathbf{G}\) requires \(O(NH^2)\) time and \(O(H^2)\) memory. In practice, this cost is incurred only during calibration and fits comfortably on modern datacenter GPUs (\eg A100-class devices) even for multi-billion-parameter LLMs, but may become restrictive on memory-constrained accelerators.
Compression involves solving a \(K \times K\) system and one GEMM, \(O(K^3 + OHK)\), with \(K \ll H\). In our experiments, a few hundreds to thousand sequences for LLMs suffice, for CNNs / ViTs we use 128 samples for calibration. Stability is ensured by setting \(\lambda = \alpha \cdot \mathrm{mean}\,\mathrm{diag}(\mathbf{G}_{PP})\) with \(\alpha \in [10^{-4}, 5 \cdot 10^{-3}]\).

\subsection{Attention blocks, compensation for LLMs}

We adopt the same producer–consumer view and Gram–regression compensation used for vision models, but decoder-only LLMs impose additional structure. The key differences concern: (i) \emph{where} activations are sampled, (ii) the \emph{head-structured} nature of self-attention (including grouped-query attention, GQA), and (iii) how pruning and folding are instantiated at the head level. Our method employs a closed-loop compensation mechanism, by re-evaluating the Gram matrix for each layer based on the output of the already-pruned previous layers, to dynamically adapt to the structural changes. This sequential alignment prevents error propagation and ensures that the regression-based compensation remains accurate throughout the entire depth of the model.

\paragraph{Where to sample activations.}
For each submodule we form the Gram matrix at the \emph{consumer input}. In pre-LN LLMs this means:
(i) MLP: post-GELU vectors $x\in\mathbb{R}^{H}$ at the input of the projection matrix $\mathbf{W}_{\mathrm{proj}}$,
(ii) Self-attention: the concatenated per-head vector $x\in\mathbb{R}^{H}$ \emph{before} the output projection $\mathbf{W}_{\mathrm{o}}$.
Over batches $n$ and time steps $t$ we accumulate the uncentered second moment
\[
\mathbf{G} \;=\; \sum_{n,t} x_{n,t}\,x_{n,t}^\top \;\in\; \mathbb{R}^{H\times H}.
\]

\paragraph{Head-structured reduction in attention.}
Let $d_{\text{model}}$ be the hidden size. Multi-head attention factorizes the attention feature axis as $H=n_{\mathrm{h}}d_{\mathrm{h}}$, where $n_{\mathrm{h}}$ is the number of heads and $d_{\mathrm{h}}$ is the per-head width. Any reduction must preserve the reshape/split invariants of attention and therefore act at the \emph{head} level. If $\mathbf{R}_{\text{heads}}\in\mathbb{R}^{K\times n_{\mathrm{h}}}$ reduces heads to $K\!\ll\! n_{\mathrm{h}}$, its action on features is the Kronecker lift
\begin{equation}
\label{eq:kron-llm}
\mathbf{R}_{\text{feat}} \;=\; \mathbf{R}_{\text{heads}} \,\otimes\, \mathbf{I}_{d_{\mathrm{h}}}
\;\in\; \mathbb{R}^{(K d_{\mathrm{h}})\times (n_{\mathrm{h}} d_{\mathrm{h}})},
\end{equation}
where $\mathbf{I}_{d_{\mathrm{h}}}$ is the $d_{\mathrm{h}}\times d_{\mathrm{h}}$ identity. For GQA with $G$ query groups and $n_{\mathrm{kv}}$ KV heads per group ($n_{\mathrm{h}}=G\,n_{\mathrm{kv}}$), the valid head reducer is block-diagonal,
\[
\mathbf{R}_{\text{blk}}
= \operatorname{blkdiag}(\underbrace{\mathbf{R}_{\mathrm{kv}},\dots,\mathbf{R}_{\mathrm{kv}}}_{G})
\in\mathbb{R}^{(G K_{\mathrm{kv}})\times (G n_{\mathrm{kv}})},
\]
\[
\mathbf{R}_{\text{feat}}=\mathbf{R}_{\text{blk}}\otimes \mathbf{I}_{d_{\mathrm{h}}}.
\]
These constraints are specific to LLM attention and typically absent in ViT MLP/channel reductions.

\paragraph{Pruning vs.\ folding at the head level.}
In attention, width reduction must act at the level of heads.  Pruning selects a subset $P \subset \{1,\dots,n_h\}$ with $|P| = K$ and uses the selection matrix $\mathbf{S} \in \{0,1\}^{K \times n_h}$, whose Kronecker lift $\mathbf{R}_{\text{feat}} = \mathbf{S} \otimes \mathbf{I}_{d_h}$ operates on the concatenated feature axis. Folding instead mixes heads via a centroid map $\mathbf{R}_{\text{heads}}$ whose rows sum to one, and lifts it as in~\Eqref{eq:kron-llm}, $\mathbf{R}_{\text{feat}} = \mathbf{R}_{\text{heads}} \otimes \mathbf{I}_{d_h}$. In this case, the reduced feature is a mixture of all heads, so there is no submatrix $\mathbf{H}_P$ of kept activations, compensation must therefore work with the full mixer $\mathbf{R}_{\text{feat}}$, not column selection.

\section{Experimental Results}
\label{sec:experiments}

Our experiments address two core questions about the performance of \name:
\begin{enumerate}
    \item Does \name consistently improve performance across architectures (ResNets, ViTs, LLMs), datasets (CIFAR-10, ImageNet-1K, C4 / WikiText-2 / PTB), and compression regimes (pruning, folding)? How does it compare to existing post-hoc compensation methods?
    \item How sensitive is \name to calibration-set size, and what are its practical computational and memory overheads?
\end{enumerate}

\subsection{\name performance on ResNet-18, ViT-B/32 and CLIP ViT-B/32}

\begin{figure}[t]
    \centering

    \begin{minipage}[t]{0.32\linewidth}
        \centering
        \includegraphics[width=\linewidth]{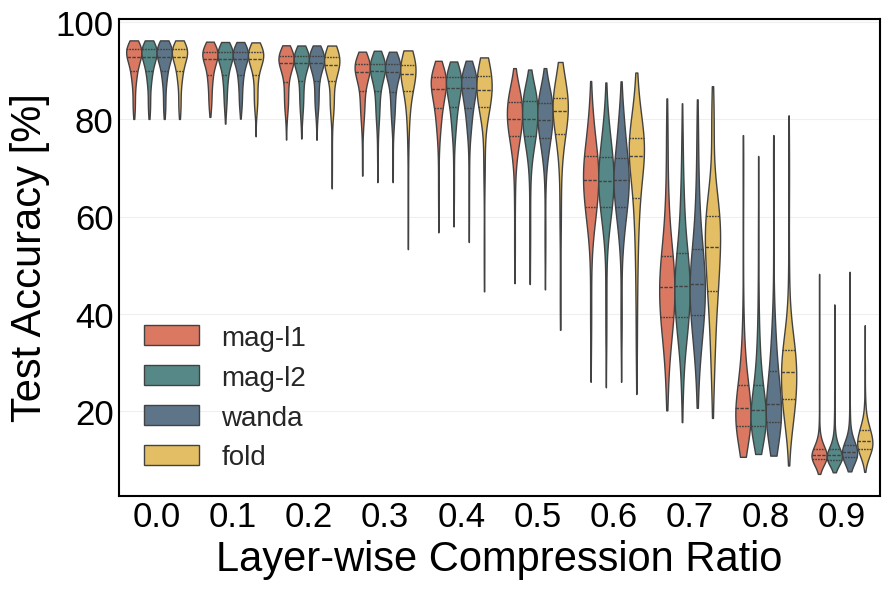}
        \caption*{(a) Test accuracy vs. layer-wise uniform compression ratio}
    \end{minipage}
    \hfill
    \begin{minipage}[t]{0.32\linewidth}
        \centering
        \includegraphics[width=\linewidth]{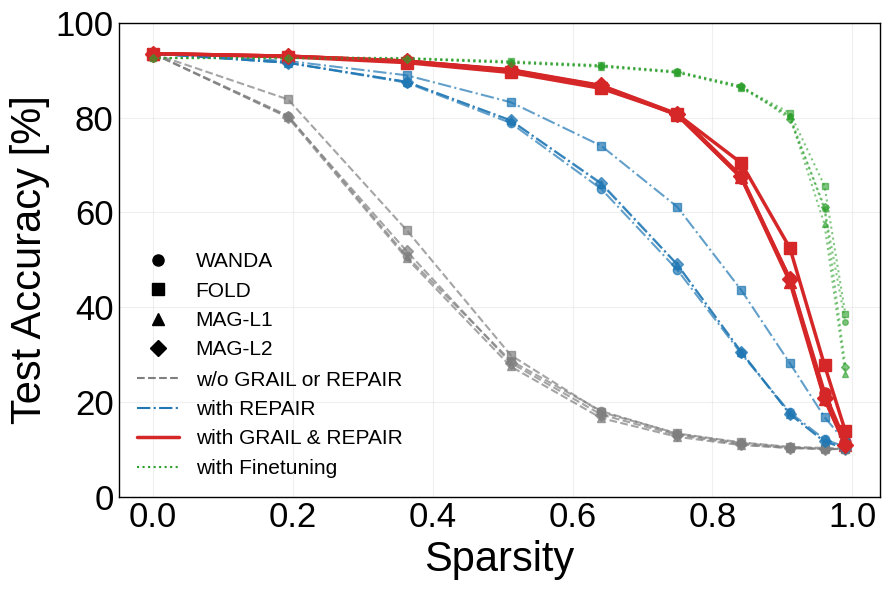}
        \caption*{(b) Mean accuracy vs. sparsity, against REPAIR and finetuning}
        \label{fig:resnet_acc}
    \end{minipage}
    \hfill
    \begin{minipage}[t]{0.32\linewidth}
        \centering
        \includegraphics[width=\linewidth]{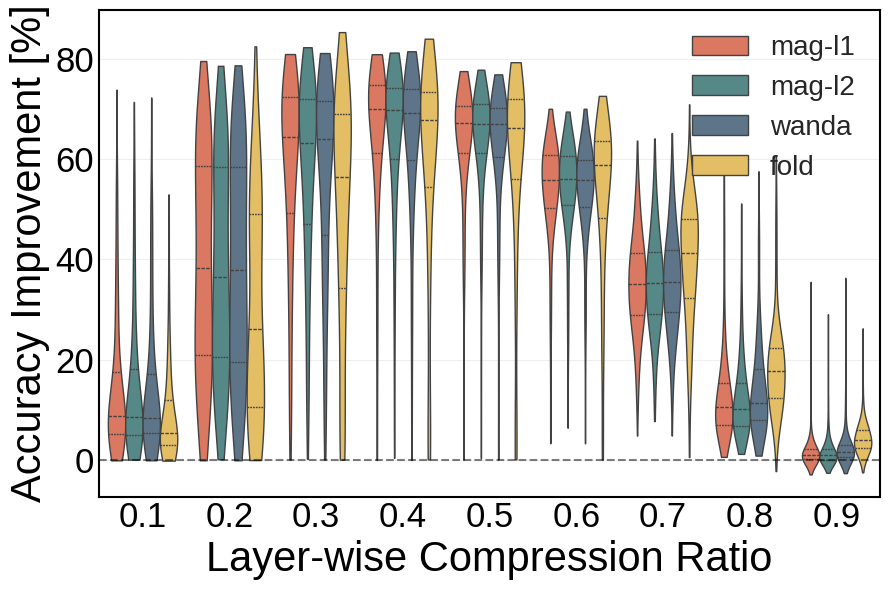}
        \caption*{(c) Relative improvement from compensation by \name}
    \end{minipage}

    \caption{\textbf{\name on 576 SGD-trained ResNet-18 models on CIFAR-10.} Using checkpoints from \citet{saukh2025cutlessfoldmore}, we find that \name consistently improves both pruning (most notably at low compression ratios) and folding (at moderate to high ratios), as shown in panels (a) and (c). It also outperforms REPAIR~\citep{jordan2023repairrenormalizingpermutedactivations}, further narrowing the gap to models fine-tuned for 5 epochs (panel (b)).}
    \label{fig:resnet_comp}
\end{figure}

We evaluate the effectiveness of \name across diverse architectures (ResNet, ViT, and CLIP), training regimes, datasets (CIFAR-10 and ImageNet-1K), and compression settings. Our study uses a large collection of 773 pretrained checkpoints to ensure robustness and generality: 576 SGD-trained ResNet-18 models on CIFAR-10 from \citet{saukh2025cutlessfoldmore}, 125 ViT-B/32 models from \citet{andriushchenko2023modernlookrelationshipsharpness}, and 72 CLIP ViT-B/32 checkpoints on ImageNet-1K from \citet{wortsman2022modelsoupsaveragingweights}.

\begin{figure}[t]
    \centering

    \begin{minipage}[t]{0.32\linewidth}
        \centering
        \includegraphics[width=\linewidth]{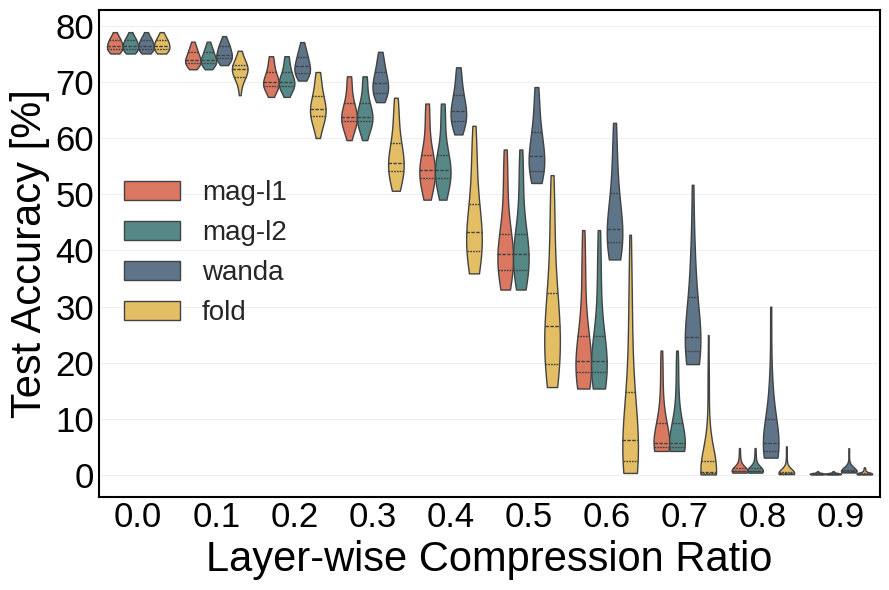}
        \caption*{(a) Test accuracy vs. layer-wise uniform compression ratio}
    \end{minipage}
    \hfill
    \begin{minipage}[t]{0.32\linewidth}
        \centering
        \includegraphics[width=\linewidth]{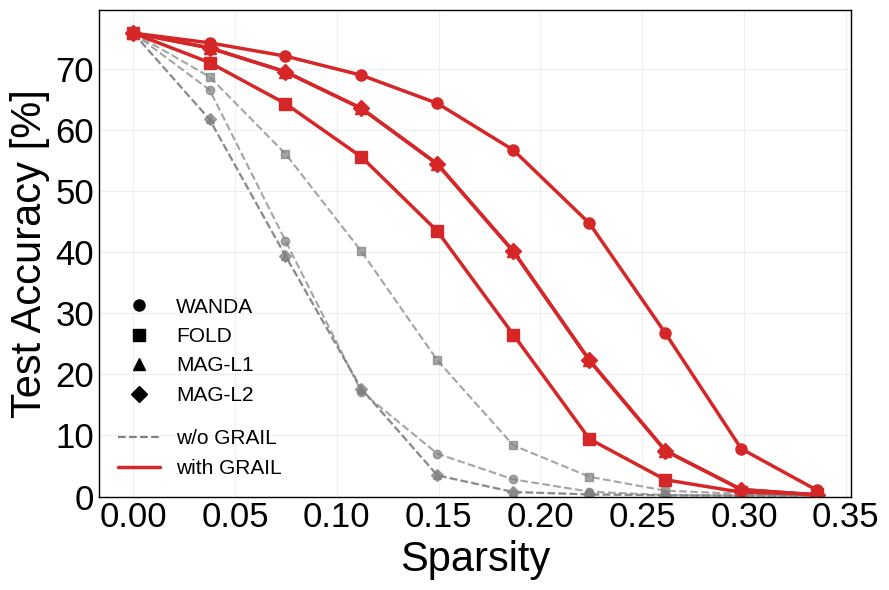}
        \caption*{(b) Mean accuracy vs. sparsity }
    \end{minipage}
    \hfill
    \begin{minipage}[t]{0.32\linewidth}
        \centering
        \includegraphics[width=\linewidth]{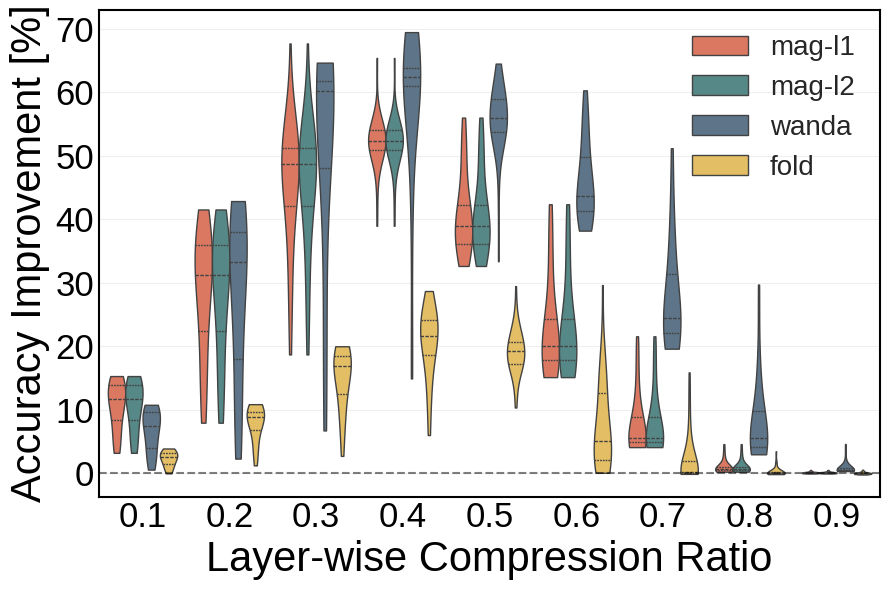}
        \caption*{(c) Relative improvement from compensation by \name}
    \end{minipage}

    \caption{\textbf{\name on 72 CLIP ViT-B/32 checkpoints on ImageNet-1K.} Using checkpoints from \citet{wortsman2022modelsoupsaveragingweights}, we find that \name consistently improves both pruning and folding (panels (a) and (c)), with pruning benefiting more strongly for all compression ratios. Consequently, \name-compensated folded models trail their \name-compensated pruned counterparts.}
    \label{fig:clip_comp}
\end{figure}

We apply our framework to four structured width-reduction methods: magnitude pruning (L1 and L2), structured Wanda, and model folding, under uniform layer-wise compression ratios from 0.1 to 0.9. For each compressed model, we compare accuracy with and without our compensation step, and also evaluate its combination with BatchNorm REPAIR~\citep{jordan2023repairrenormalizingpermutedactivations}. Although \name is data-aware, it remains entirely training-free: calibration uses only 128 unlabeled images for vision models, and all steps require no gradients.

We evaluate the effectiveness of our proposed compensation framework across various compression ratios, ranging from 10\% to 90\% layer-wise compression ratio. \Figref{fig:resnet_comp}, \Figref{fig:clip_comp} and \Figref{fig:vit_comp} summarize the results for ResNet-18 on CIFAR-10, CLIP ViT-B/32 on ImageNet-1K and ViT-B/32 on CIFAR-10 (in appendix) under different compression paradigms (magnitude pruning, Wanda, and model folding). \name achieves a significant accuracy recovery from the compressed model. At low to moderate sparsity (10\%-40\%), the compensated models achieve near-lossless compression, often recovering the accuracy up to within 0.5\% of the original model. In some cases, \eg ResNet-18 at 20\% sparsity, the compensated model even slightly surpasses the baseline, suggesting a regularization effect. At moderate to high sparsity (50\%-80\%), the advantage of \name becomes most pronounced. For instance, at 65\% sparsity using L1-magnitude pruning, the baseline accuracy of ResNet-18 collapses to 17.6\%, whereas \name restores it to 84.8\%, a remarkable 67.2\% improvement. Similarly, for ViT and CLIP models, which are known to be sensitive to structured pruning, our framework effectively mitigates the catastrophic accuracy drop typically observed at high compression rates. Our approach proves to be agnostic to the underlying compression operator, validating its universality as a plug-and-play module for post-training compression and compensation.

\subsection{\name performance on LLaMA-2-7B}
\label{sec:llm_results}

We evaluate \name on LLaMA-2-7B across three standard language modeling benchmarks: C4, WikiText-2, and PTB. We consider sparsity levels from 10\% to 70\% and apply five representative structured pruning methods: ZipLM, structured Wanda, Wanda++, SlimGPT, and FLAP. ZipLM can be regarded as a structured variant of SparseGPT. Since pruning and weight updates are jointly performed and cannot be decoupled, \name is not applicable to ZipLM. All methods use 128 calibration sequences of length 2048, and performance is reported in terms of test perplexity.

\begin{table*}[h]
\centering
\caption{\textbf{Perplexity ($\downarrow$) comparison on LLaMA-2-7B} under different sparsity levels across three datasets (C4, PTB, WikiText2) with 2048
sequence length and 128 calibration samples.}
\label{tab:wc_compare_all}
\setlength{\tabcolsep}{4pt}
\renewcommand{\arraystretch}{1.0}

\begin{tabular}{c|c|ccccccc}
\hline
\textbf{Dataset} & \textbf{Method} 
& \textbf{10\%} & \textbf{20\%} & \textbf{30\%} 
& \textbf{40\%} & \textbf{50\%} & \textbf{60\%} & \textbf{70\%} \\
\hline

\rowcolor{blue!10}
C4 & ZipLM & 8.26 & 9.86 & 12.26 & 16.78 & \hi{21.28} & \hi{35.76} & 2291.52 \\
& Wanda & 8.25 & 9.88 & 12.49 & 20.80 & 131.54 & 155.41 & 508.51 \\
\rowcolor{blue!10} & Wanda \hi{+ \name}   & \hi{8.05} & \hi{9.28} & \hi{11.26} & 15.50 & 25.97 & 213.92 & 2127.03 \\
& Wanda++   & 8.07 & 9.32 & 11.74 & 15.89 & 26.72 & 77.35 & 132.58 \\
\rowcolor{blue!10} & Wanda++ \hi{+ \name}   & \hi{8.05} & \hi{9.28} & \hi{11.26} & \hi{15.05} & 23.75 & 187.21 & 1688.64 \\
& SlimGPT & 11.41 & 18.80 & 23.51 & 57.75 & 235.56 & 1288.51 & 4049.08 \\
\rowcolor{blue!10} & SlimGPT \hi{+ \name}   & 8.21 & 9.49 & 11.66 & 15.83 & 28.05 & 67.63 & \hi{102.46} \\
              
& FLAP  & 8.35 & 10.26 & 12.84 & 17.41 & 22.94 & 100.53 & 803.27 \\
\rowcolor{blue!10} & FLAP \hi{+ \name}    & 8.26 & 10.05 & 12.40 & 16.45 & 22.04 & 40.77 & 126.88 \\
\hline

\rowcolor{blue!10}
PTB & ZipLM & 26.87 & 32.39 & 40.21 & 139.20 & 179.51 & 155.00 & 268.90 \\
& Wanda & 22.01 & 30.91 & 80.70 & 174.46 & 287.26 & 423.63 & 840.68 \\
\rowcolor{blue!10} & Wanda \hi{+ \name}   & 21.85 & 24.31 & 25.94 & 49.25 & 62.86 & 425.26 & \hi{215.91} \\
& Wanda++ & 22.14 & 26.32 & 30.95 & 37.85 & 54.29 & 90.80 & 338.67 \\ 
\rowcolor{blue!10} & Wanda++ \hi{+ \name}   & 22.37 & 25.21 & 25.85 & 85.64 & 56.49 & 592.58 & 217.58 \\
& SlimGPT & 37.80 & 56.94 & 85.82 & 419.73 & 938.52 & 3683.63 & 7983.82 \\
\rowcolor{blue!10} & SlimGPT \hi{+ \name}   & 22.68 & 25.74 & 35.99 & 39.47 & 46.62 & 76.58 & 217.83 \\

& FLAP & 22.38 & 24.99 & 29.22 & 37.34 & 54.34 & 100.01 & 1157.63 \\
\rowcolor{blue!10} & FLAP \hi{+ \name}    & \hi{21.63} & \hi{23.05} & \hi{25.01} & \hi{28.90} & \hi{35.58} & \hi{52.16} & 576.46 \\
\hline

\rowcolor{blue!10}
WikiText2 & ZipLM & 5.84 & 6.36 & 7.64 & nan & 12.65 & 21.11 & 45.74 \\
& Wanda & 6.18 & 7.45 & 9.18 & 15.16 & 171.29 & 272.47 & 1839.20 \\
\rowcolor{blue!10} & Wanda \hi{+ \name}   & \hi{5.75} & 6.44 & 7.45 & 9.98  & 18.85 & 39.59 & 408.68 \\
& Wanda++ & 5.80 & 6.56 & 7.59 & 10.18 & 23.29 & 44.00 & 128.00 \\
\rowcolor{blue!10} & Wanda++ \hi{+ \name} & \hi{5.75} & 6.45 & 7.44 & 10.44 & 16.14 & 35.17 & 288.79 \\
& SlimGPT & 7.69 & 9.81 & nan & 62.56 & 590.75 & 1220.71 & 16764.33 \\
\rowcolor{blue!10} & SlimGPT \hi{+ \name}   & 5.81 & \hi{6.32} & \hi{7.34} & \hi{9.71}  & 16.30 & 23.45 & \hi{43.14} \\              
& FLAP & 6.01 & 7.16 & 8.85 & 11.49 & 16.67 & 31.80 & 490.85 \\
\rowcolor{blue!10} & FLAP \hi{+ \name}    & 5.88 & 6.80 & 8.08 & 10.18 & \hi{13.45} & \hi{20.46} & 71.63 \\
\hline

\end{tabular}
\end{table*}

Table~\ref{tab:wc_compare_all} summarizes the results. Across all datasets and sparsity levels, \name helps to considerably reduce perplexity relative to the corresponding pruning baselines. The benefit of \name is most pronounced for methods that perform aggressive structured removal without explicit activation reconstruction. For instance, SlimGPT exhibits rapid perplexity growth beyond 30\% sparsity, whereas \name substantially stabilizes its behavior across all three datasets. FLAP, which already includes bias compensation, also benefits from \name at higher sparsities, indicating that correcting second-order activation geometry complements existing first-order compensation schemes. We further compare \name to regional optimization  used in Wanda++. Regional optimization performs local gradient-based fine-tuning of retained weights, whereas \name applies a closed-form, deterministic reconstruction. As shown in Table~\ref{tab:wc_compare_all}, \name achieves comparable or better perplexity recovery for low sparsity levels without requiring gradients or iterative optimization. 

Table~\ref{tab:zs_compare} reports zero-shot accuracies for five structured pruning baselines—ZipLM, FLAP, Wanda, SlimGPT, and Wanda++ at 20\% and 50\% sparsity, evaluated both with and without our \name compensation, except for ZipLM. Across all pruning methods and benchmarks, \name often improves or preserves zero-shot accuracy, confirming that reconstruction of pruned feature directions is broadly beneficial and algorithm-agnostic. For some datasets, such as ARC-E and BoolQ, the improvements are consistent and significant across all compression methods and sparsity levels. These results highlight that \name provides a simple, training-free mechanism to recover destruction of cross-channel geometry, yielding benefits across different pruning criteria and task types.

\begin{table*}[h]
\centering
\caption{\textbf{Zero-shot accuracy ($\uparrow$) comparison on LLaMA-2-7B} for different model compression methods with and without \name compensation calibrated on C4 dataset across six benchmarks.}
\label{tab:zs_compare}
\setlength{\tabcolsep}{5pt}
\renewcommand{\arraystretch}{1.1}

\begin{tabular}{c|c|cccccc}
\hline
\textbf{Sparsity} & \textbf{Method} 
& \textbf{ARC-C} 
& \textbf{ARC-E} 
& \textbf{HellaSwag} 
& \textbf{PIQA} 
& \textbf{BoolQ} 
& \textbf{Winogrande} \\
\hline

\rowcolor{blue!10}
20\% & ZipLM & 0.3763 & 0.7083 & 0.5116 & 0.7388 & 0.5697 & 0.6685  \\
& Wanda  
  & 0.3865 & 0.7096 & \hi{0.5452} & 0.7633 & 0.7141 & 0.6417 \\

\rowcolor{blue!10} & Wanda \hi{+ \name}      
  & \hi{0.4164} & \hi{0.7382} & 0.5348 & 0.7655 & 0.7416 & \hi{0.6843} \\
  
& Wanda++
  & 0.3959 & 0.7256 & 0.5436 & \hi{0.7688} & 0.7083 & 0.6827 \\

\rowcolor{blue!10} & Wanda++ \hi{+ \name}      
  & 0.4104 & 0.7357 & 0.5357 & 0.7644 & \hi{0.7446} & 0.6772 \\
  
& SlimGPT 
  & 0.3200 & 0.5829 & 0.4781 & 0.7133 & 0.6642 & 0.5359 \\

\rowcolor{blue!10} & SlimGPT \hi{+ \name}      
  & 0.4138 & 0.7256 & 0.5373 & 0.7628 & 0.7431 & 0.6819 \\
  
& FLAP 
  & 0.3609 & 0.6759 & 0.5170 & 0.7508 & 0.6872 & 0.6638 \\

\rowcolor{blue!10} & FLAP \hi{+ \name}    
  & 0.3490 & 0.6776 & 0.4952 & 0.7503 & 0.7086 & 0.6606 \\

\hline

\rowcolor{blue!10}
50\% & ZipLM & 0.2423 & 0.4465 & 0.3469 & 0.6023 & 0.5865 & 0.5564  \\

& Wanda 
  & 0.1937 & 0.2795 & 0.2680 & 0.5424 & 0.5911 & 0.4838 \\

\rowcolor{blue!10} & Wanda \hi{+ \name}   
  & 0.2082 & 0.4630 & 0.3379 & 0.6485 & 0.6144 & 0.5446 \\
  
& Wanda++ 
  & 0.2031 & 0.3636 & 0.3140 & 0.6072 & 0.6254& 0.5012 \\

\rowcolor{blue!10} & Wanda++ \hi{+ \name}      
  & 0.2133 & 0.4638 & 0.3385 & 0.6480 & 0.6153 & 0.5280 \\

& SlimGPT  
  & 0.2116 & 0.2917 & 0.2697 & 0.5533 & 0.5575 & 0.5296 \\

\rowcolor{blue!10} & SlimGPT \hi{+ \name}   
  & 0.2406 & 0.4731 & 0.3755 & \hi{0.6638} & 0.6349 & 0.5864 \\
  
& FLAP 
  & \hi{0.2679} & 0.4718 & \hi{0.3833} & 0.6529 & \hi{0.6471} & \hi{0.5983} \\

\rowcolor{blue!10} & FLAP \hi{+ \name}     
  & 0.2449 & \hi{0.4819} & 0.3696 & 0.6551 & 0.6260 & \hi{0.5983} \\

\hline

\end{tabular}
\end{table*}

\subsection{Data and resource efficiency}

\begin{figure*}[t]
\centering
\begin{minipage}[t]{0.66\linewidth}
    \centering
    \begin{subfigure}{0.49\linewidth}
        \centering
        \includegraphics[width=\linewidth]{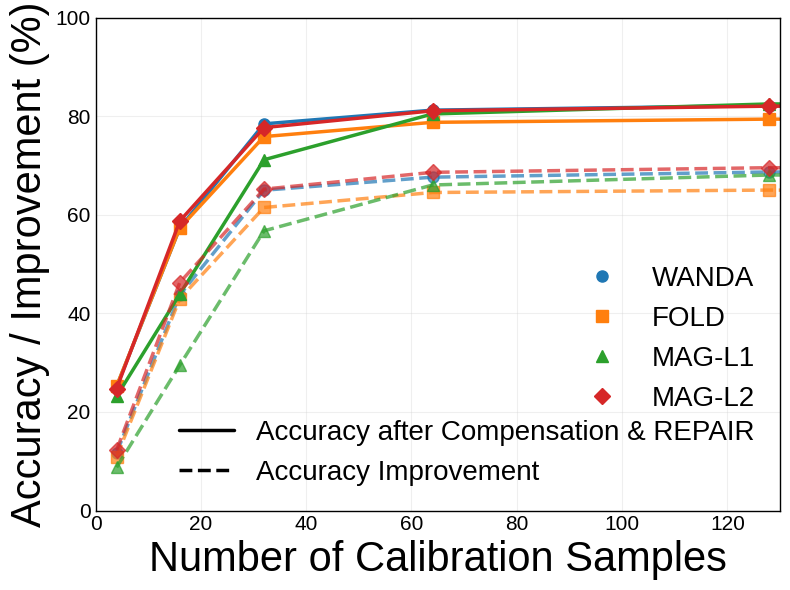}
        \caption{ResNet-18 (75\% sparsity)}
        \label{fig:calibration_size_resnet}
    \end{subfigure}\hfill
    \begin{subfigure}{0.49\linewidth}
        \centering
        \includegraphics[width=\linewidth]{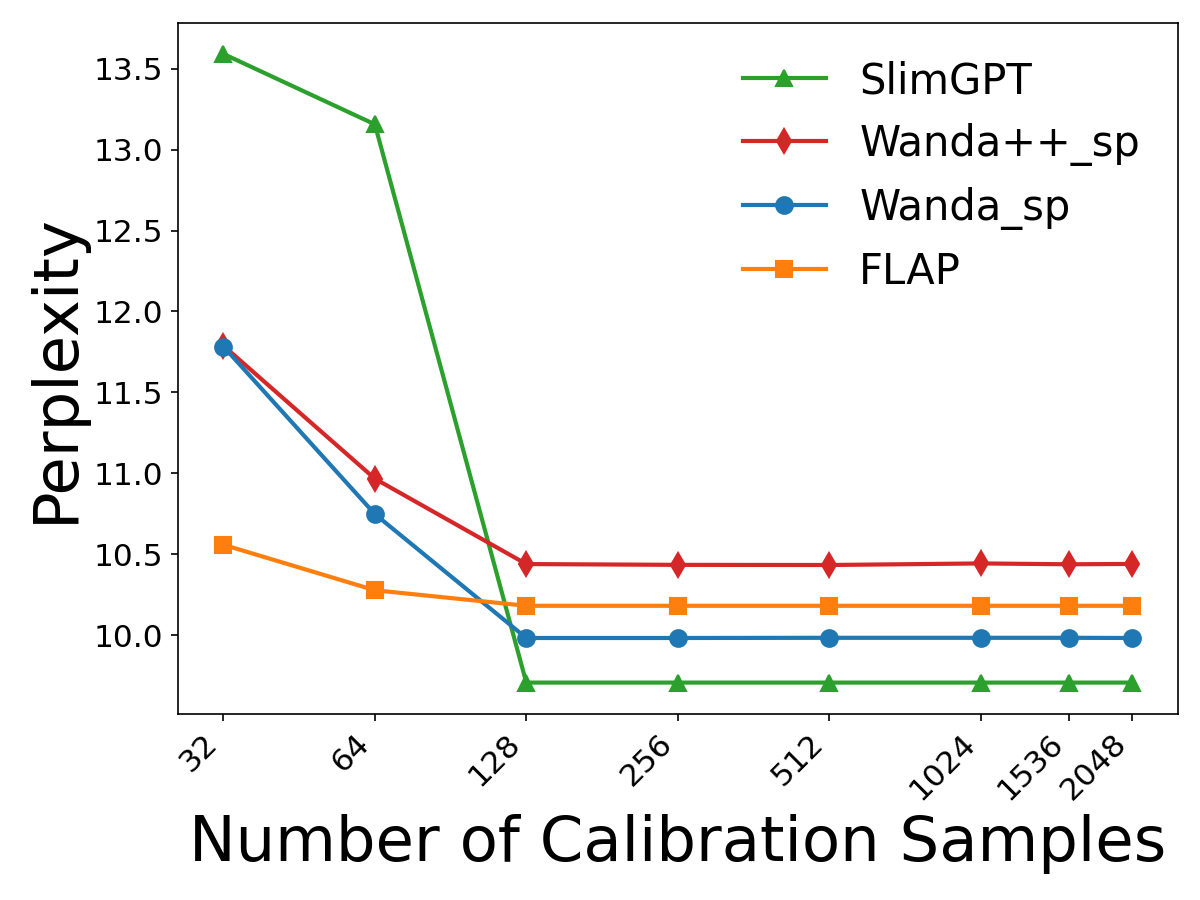}
        \caption{LLaMA-2-7B (40\% sparsity)}
        \label{fig:calibration_size_llm}
    \end{subfigure}
\end{minipage}
\hfill
\begin{minipage}[t]{0.33\linewidth}
    \vskip -4.4cm
    \caption{
        \textbf{Ablation on compensation dataset size.} 
        \textbf{Left:} effect on accuracy recovery for ResNet-18 on CIFAR-10 at 75\% sparsity. 
        \textbf{Right:} effect on LLaMA-2-7B perplexity at 40\% sparsity on WikiText-2. In our experiments we use 128 unlabeled images for vision models (ResNet-18, ViT, CLIP) and only 128 sequences of length 2048 tokens for LLMs.  
    }
    \label{fig:calibration_size}
\end{minipage}
\end{figure*}

To evaluate the data efficiency of \name, \Figref{fig:calibration_size} illustrates the accuracy improvement—defined as the margin between the before and after weight compensation model accuracy across varying numbers of calibration samples for different compression methods on ResNet-18 with 75\% sparsity and on LLaMa-2-7B at 40\% sparsity. ResNet-18 experiments are performed over 100 checkpoints with original accuracy above 90\%, and we report the averaged performance. The results show that \name performance follows a logarithmic growth pattern: significant accuracy recovery is achieved with a very small batch of samples, after which the performance gain rapidly saturates. This observation highlights the high data efficiency of the closed-form ridge regression solution, indicating that a small calibration set is sufficient to effectively reconstruct feature maps and recover model performance. For LLaMa-2-7B, only 128 sequences of length 2048 tokens were used in our evaluation. Across all methods, calibration with at least 128 samples is sufficient to reach the performance plateau; SlimGPT requires the most samples to stabilize, whereas FLAP and Wanda exhibit stronger data efficiency.

\begin{table}[h]
    \centering
    \caption{\textbf{Overhead of \name compensation on vision and language models.} The calibration step collect activation statistics, while the compensation step comprises \name computations.}
    \label{tab:overhead_vision_models}
    \setlength{\tabcolsep}{5pt}
    \renewcommand{\arraystretch}{1.15}
    
    \begin{tabular}{l|cc|cc}
    \hline
    \textbf{Model} &
    \multicolumn{2}{c|}{\textbf{Overhead Time (s)}} &
    \multicolumn{2}{c}{\textbf{Overhead Memory (MB)}} \\
    & Calibration & Compensation & Calibration & Compensation \\
    \hline

    ResNet & 0.19 & 0.10 & 162.02  & 22.00  \\
    ViT    & 0.20 & 0.04 & 161.62 & 5.76 \\
    CLIP   & 0.95 & 0.16 & 300.00 & 139.54 \\
    LLaMA-2-7B    & 58.077 & 3.16 & 1137.12 & 3367.00 \\
    \hline
    \end{tabular}
\end{table}

To assess the computational and memory overhead introduced by \name, we measure both runtime and peak memory across all architectures and present the results in Table~\ref{tab:overhead_vision_models}. The evaluation consists of two stages: (1) a single-batch calibration pass (batch size 128 images or 256 sequences) used to collect activation statistics, and (2) the subsequent \name compensation. All measurements are performed on a single NVIDIA A100 GPU. The results show that calibration dominates the total cost, while the compensation step itself is lightweight relative to the base model size. For large models such as LLaMA-2-7B, the peak memory footprint ($\approx$3 GB) is manageable within standard datacenter GPU budgets, though this overhead may be non-trivial in more memory-constrained deployment settings.

\section{Conclusions, Limitations and Outlook}

We introduce \name, a unified, training-free layer-wise, interleaved pruning-and-compensation framework that restores the input–output behavior of structured compressed (pruning or folding) blocks through a calibration-driven linear reconstruction. Across ResNets, ViTs, and decoder-only LLMs, \name consistently improves accuracy or perplexity over compressed-only baselines and complements existing recovery mechanisms, while requiring only a small unlabeled calibration set. Limitations include its reliance on a short forward pass through the uncompressed model, sensitivity to distribution shifts in activation statistics, and its block-local nature, which may constrain performance under extreme compression.
In addition, Gram-matrix accumulation introduces an $\mathcal{O}(H^2)$ memory footprint during calibration, which is practical on datacenter-class GPUs but may limit applicability on edge or tightly budgeted accelerators. Extending \name to jointly compensate multiple layers and integrating it with quantization, KV-cache compression, or task-aware calibration signals are promising future directions.

\section*{Acknowledgments}

This work was supported by the ANT project, funded by the European Union’s Horizon Europe research and innovation programme under Grant Agreement No. 101169439, and by the FFG COMET K1 Centre “Pro$^{2}$Future II” (Cognitive and Sustainable Products and Production Systems of the Future; Contract No. 911655). The results presented in this paper were obtained using the computational resources of Pro2Future GmbH, the Central IT Services of Graz University of Technology (ZID), and the Austrian Scientific Computing (ASC) infrastructure.



\bibliographystyle{unsrtnat}
\bibliography{references}


\clearpage
\appendix
\section*{Appendix}

\section{Implementation Details}
Our vision model implementation is based on \cite{saukh2025cutlessfoldmore}\footnote{\url{https://github.com/osaukh/folding_as_projection}}.
For the LLM, we build upon the open-source FLAP framework\footnote{\url{https://github.com/CASIA-LMC-Lab/FLAP}},
into which we integrate multiple structured pruning methods under a unified interface and implement the proposed \name~compensation mechanism for consistent evaluation.
All experiments were conducted on a compute cluster equipped with $8\times$ NVIDIA A100 (40\,GB RAM) GPUs.
To ensure reproducibility, all random seeds are fixed in the configuration files and execution scripts.

\section{Use of Large Language Models}
ChatGPT-3 was used to correct grammatical errors in the manuscript and to address minor formatting issues in Overleaf. Gemini-3 Pro was used to assist
with coding and debugging programming errors. All research concepts, theoretical contributions, experimental design, and the scientific content of the manuscript were fully developed by the authors.

\section{More Results}

\begin{figure}[h]
    \centering

    \begin{minipage}[t]{0.32\linewidth}
        \centering
        \includegraphics[width=\linewidth]{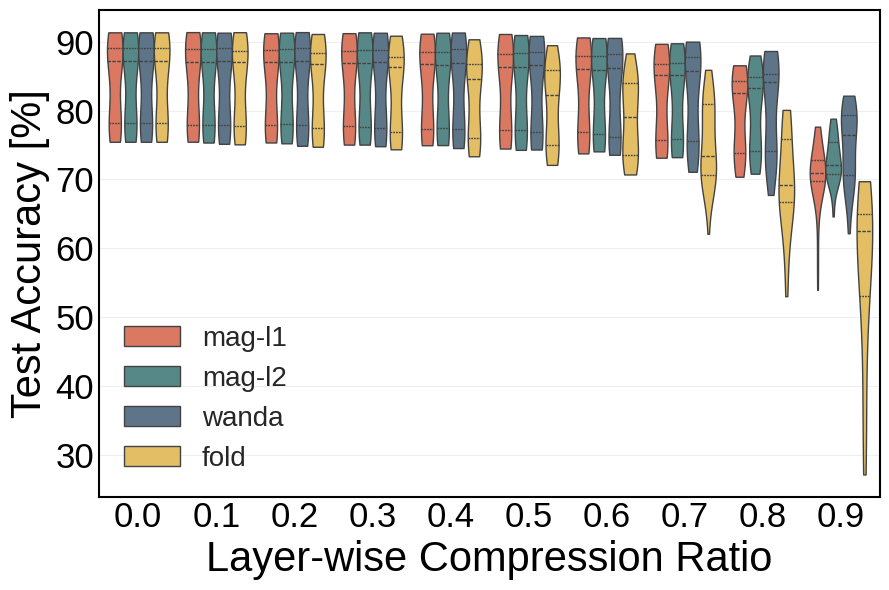}
        \caption*{(a) Test accuracy vs. layer-wise uniform compression ratio}
    \end{minipage}
    \hfill
    \begin{minipage}[t]{0.32\linewidth}
        \centering
        \includegraphics[width=\linewidth]{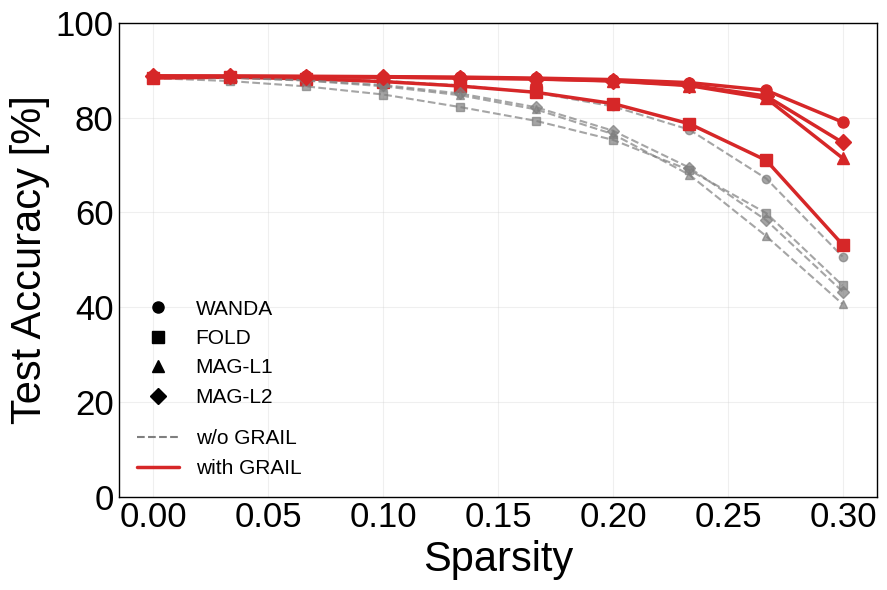}
        \caption*{(b) Mean accuracy vs. sparsity}
    \end{minipage}
    \hfill
    \begin{minipage}[t]{0.32\linewidth}
        \centering
        \includegraphics[width=\linewidth]{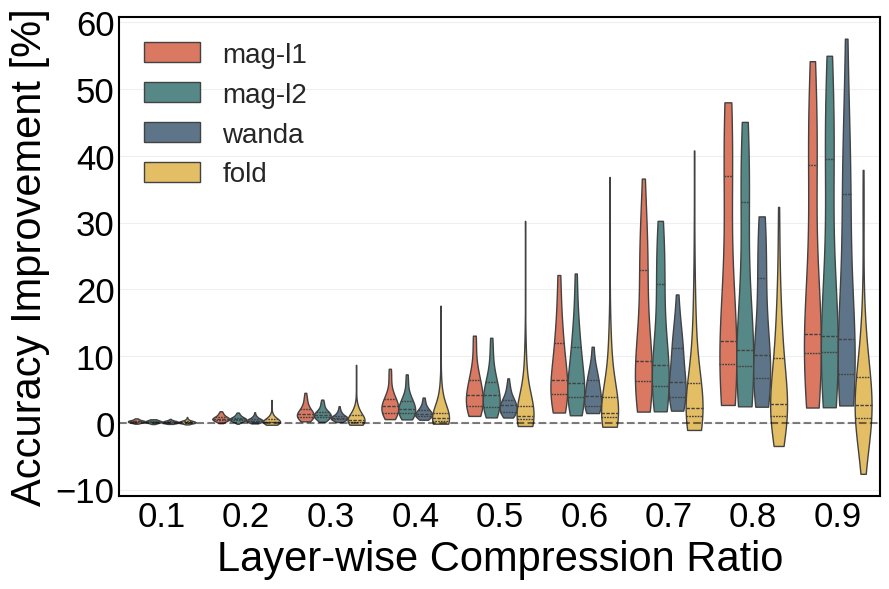}
        \caption*{(c) Relative improvement from compensation by \name}
    \end{minipage}

    \caption{\textbf{\name on 125 ViT-B/32 models on CIFAR-10.} Using checkpoints from \citet{andriushchenko2023modernlookrelationshipsharpness}, we find that \name consistently improves pruning and in most cases also folding, as shown in panels (a) and (c). However, similarly to CLIP ViT-B/32 results reported in the main paper, \name-compensated folded models lag behind \name-compensated pruned models.}
    \label{fig:vit_comp}
\end{figure}

\begin{figure*}[h]
    \centering
    
    \begin{subfigure}{0.49\textwidth}
        \centering
   
        \begin{subfigure}{0.49\textwidth}
            \centering
            \includegraphics[width=\linewidth]{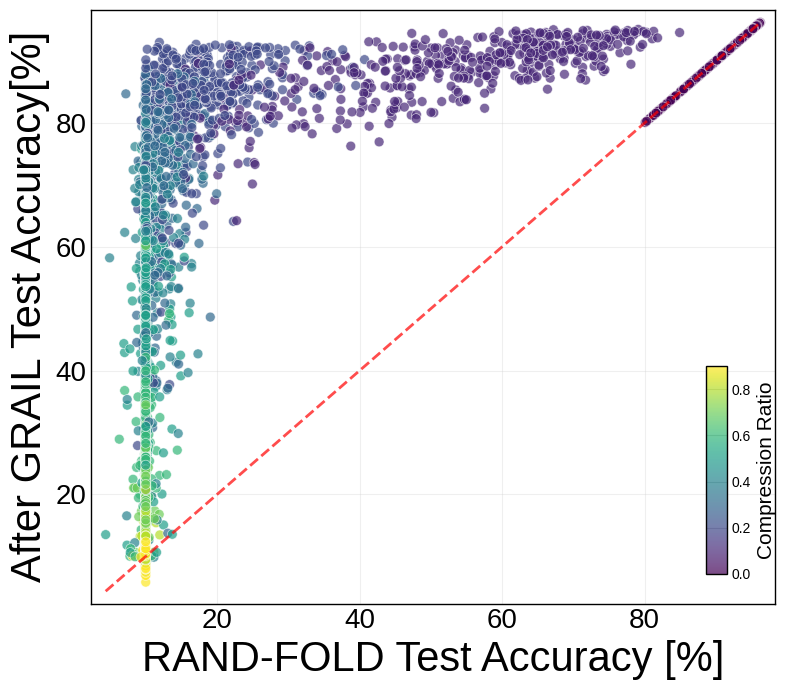}
        \end{subfigure}
        \begin{subfigure}{0.49\textwidth}
            \centering
            \includegraphics[width=\linewidth]{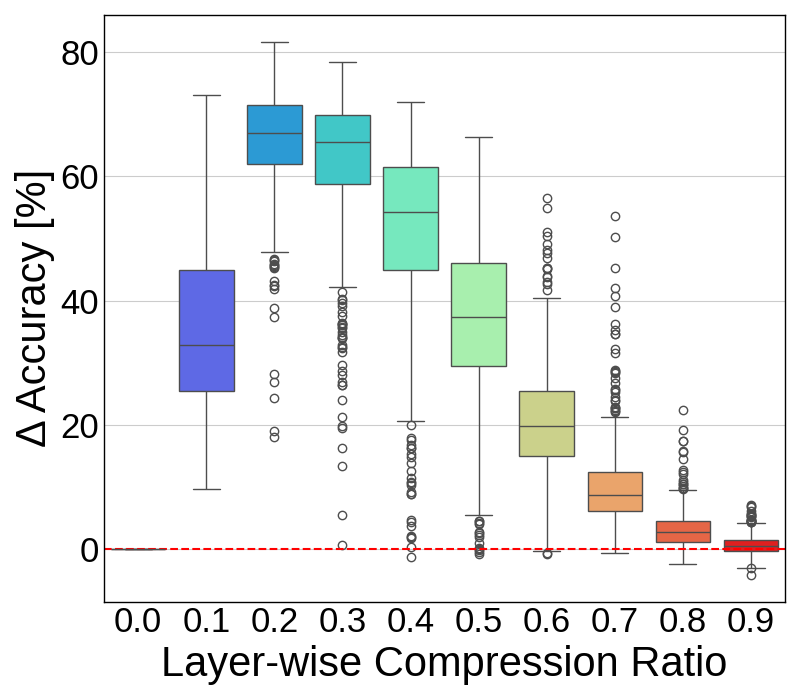}
        \end{subfigure}
        \caption{ResNet-18: Random folding, w/ vs. w/o \name}
        \label{fig:resnet_rand_fold}
    \end{subfigure}
    \hfill
     \begin{subfigure}{0.49\textwidth}
        \centering
   
        \begin{subfigure}{0.49\textwidth}
            \centering
            \includegraphics[width=\linewidth]{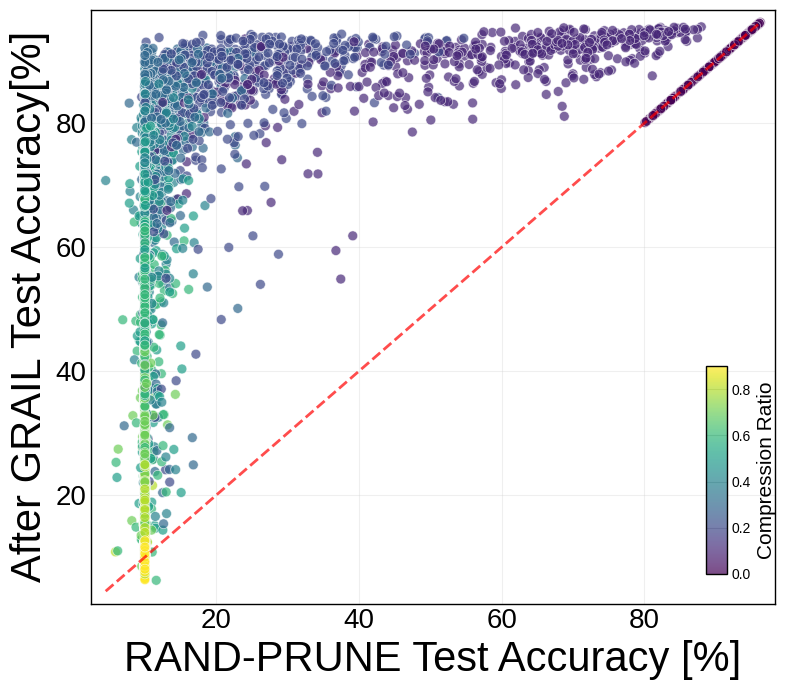}
        \end{subfigure}
        \begin{subfigure}{0.49\textwidth}
            \centering
            \includegraphics[width=\linewidth]{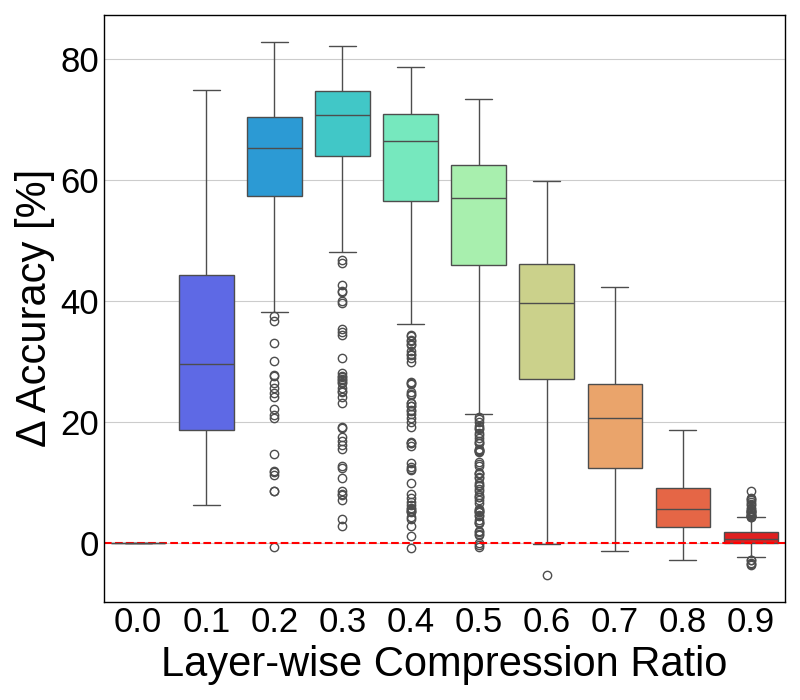}
        \end{subfigure}
        \caption{ResNet-18: Random pruning, w/ vs. w/o \name}
        \label{fig:resnet_rand_prune}
    \end{subfigure}
    \vspace{0.5em}
    
       \begin{subfigure}{0.49\textwidth}
        \centering
   
        \begin{subfigure}{0.49\textwidth}
            \centering
            \includegraphics[width=\linewidth]{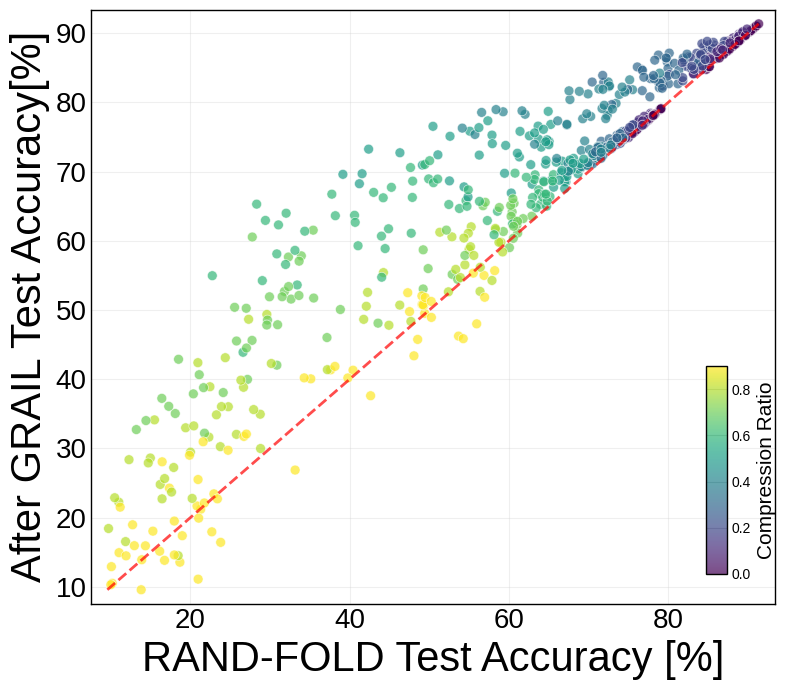}
        \end{subfigure}
        \begin{subfigure}{0.49\textwidth}
            \centering
            \includegraphics[width=\linewidth]{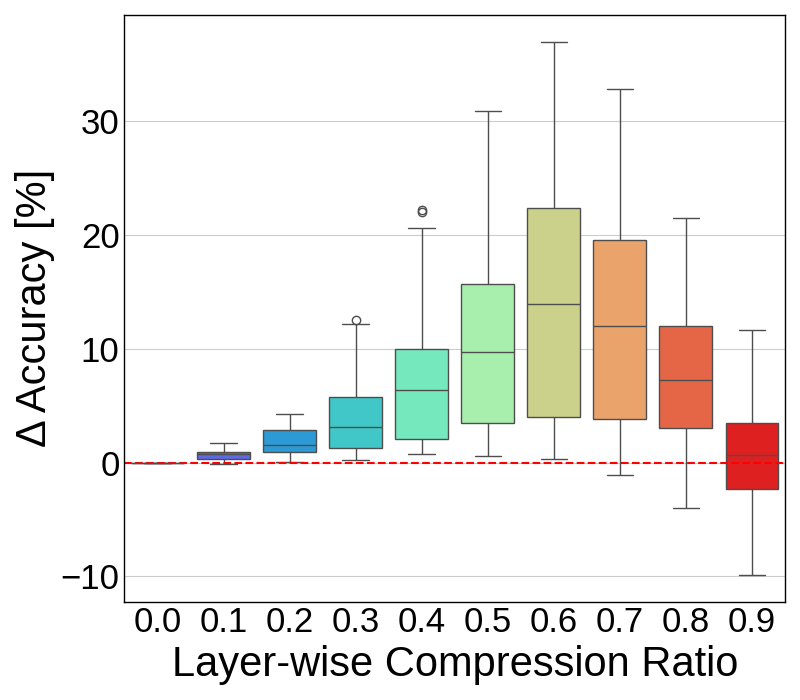}
        \end{subfigure}
        \caption{ViT-B/32: Random folding, w/ vs. w/o \name}
        \label{fig:vit_rand_fold}
    \end{subfigure}
    \hfill
     \begin{subfigure}{0.49\textwidth}
        \centering
   
        \begin{subfigure}{0.49\textwidth}
            \centering
            \includegraphics[width=\linewidth]{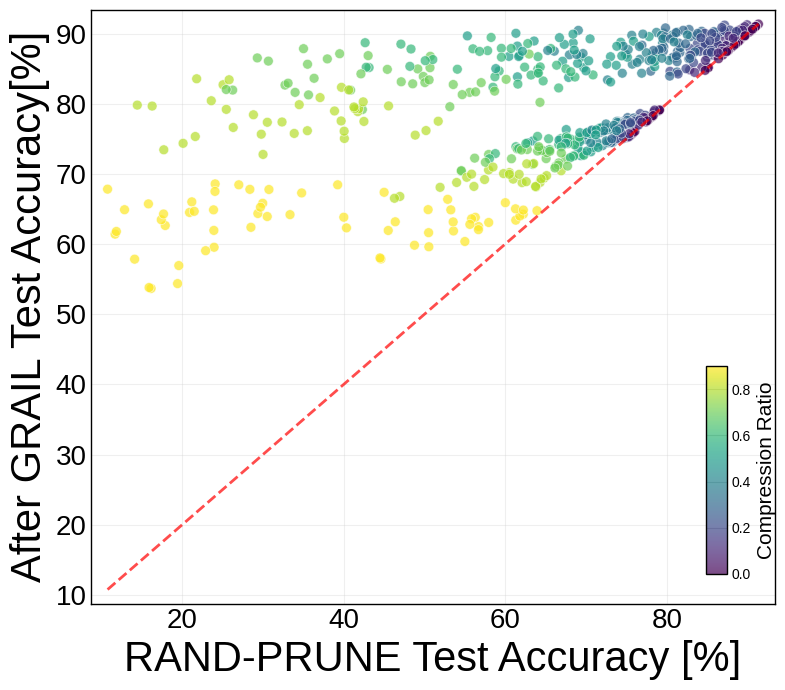}
        \end{subfigure}
        \begin{subfigure}{0.49\textwidth}
            \centering
            \includegraphics[width=\linewidth]{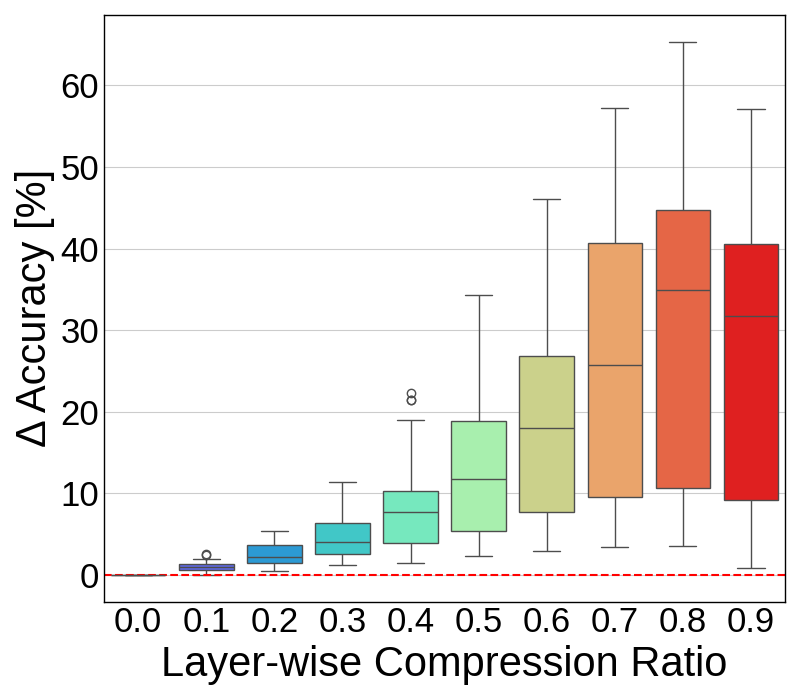}
        \end{subfigure}
        \caption{ViT-B/32: Random pruning, w/ vs. w/o \name}
        \label{fig:resnet_magl1}
    \end{subfigure}
    
    \caption{\textbf{\name on ResNet-18 and ViT-B/32 under random folding and pruning.} Across both architectures, \name consistently improves the accuracy of compressed models, as seen in the before/after scatter plots (left) and the accuracy gains across compression ratios (right) for all four settings: ResNet-18 folding (a), ResNet-18 pruning (b), ViT-B/32 folding (c), and ViT-B/32 pruning (d).}
    \label{fig:resnet_results}
\end{figure*}

\begin{figure*}[h]
    \centering

    \begin{minipage}{\textwidth}
        \centering
        \includegraphics[width=0.24\linewidth]{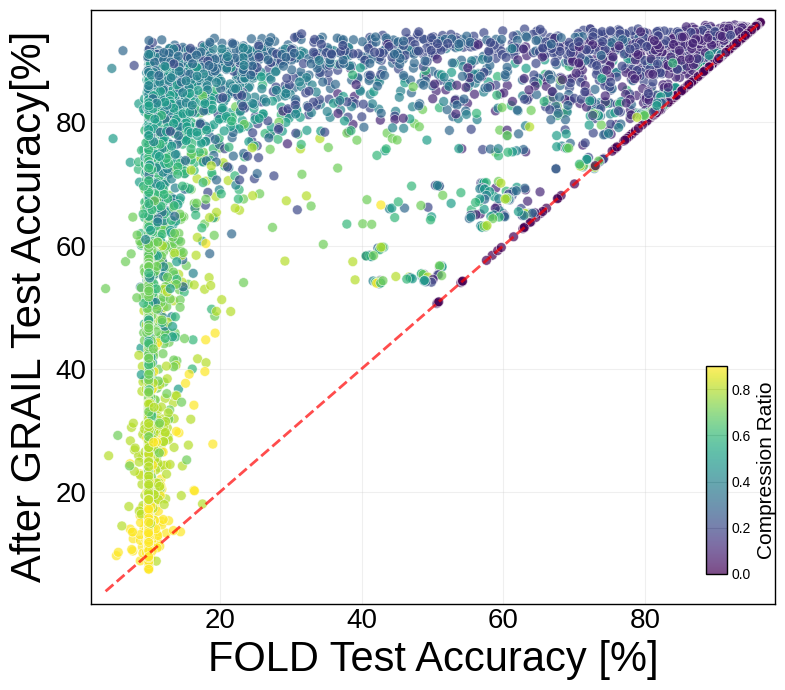}
        \includegraphics[width=0.24\linewidth]{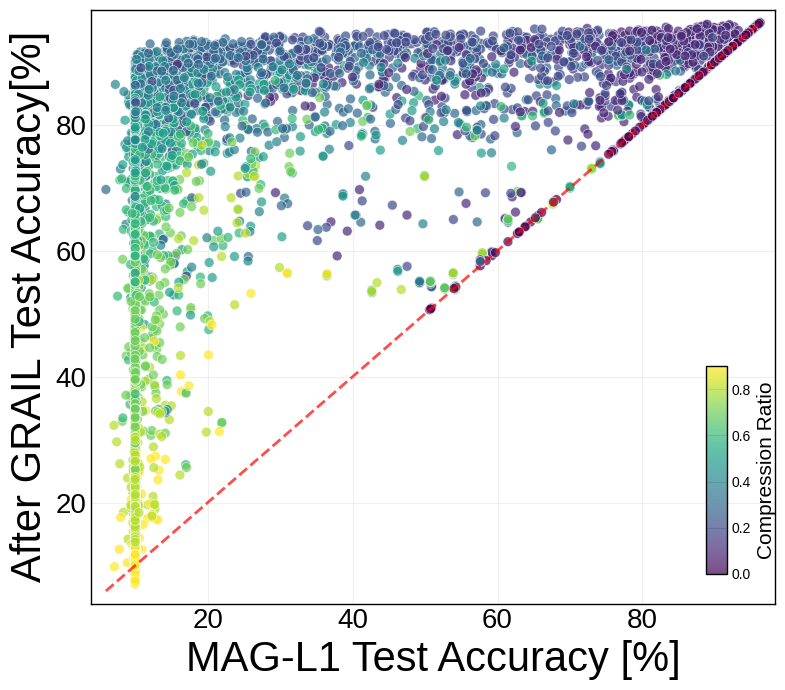}
        \includegraphics[width=0.24\linewidth]{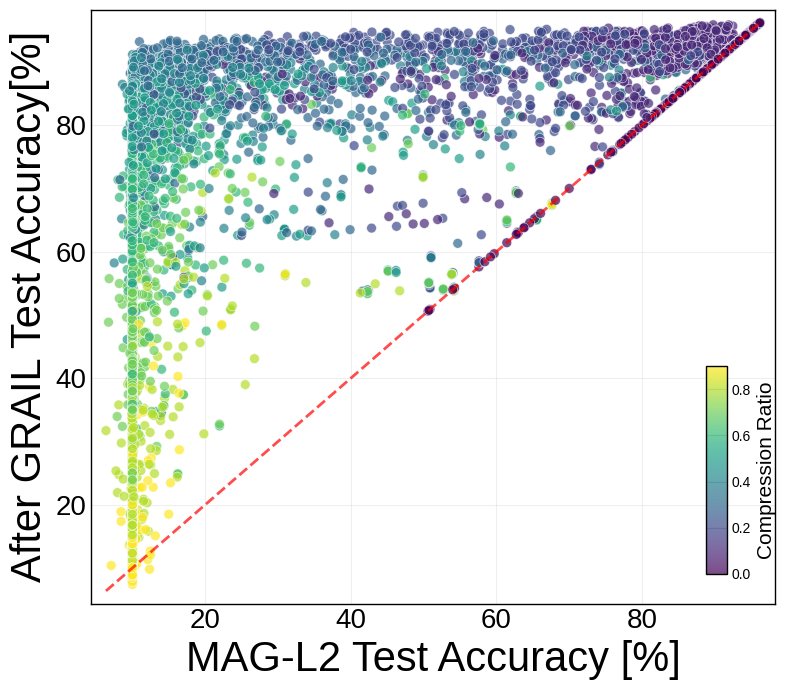}
        \includegraphics[width=0.24\linewidth]{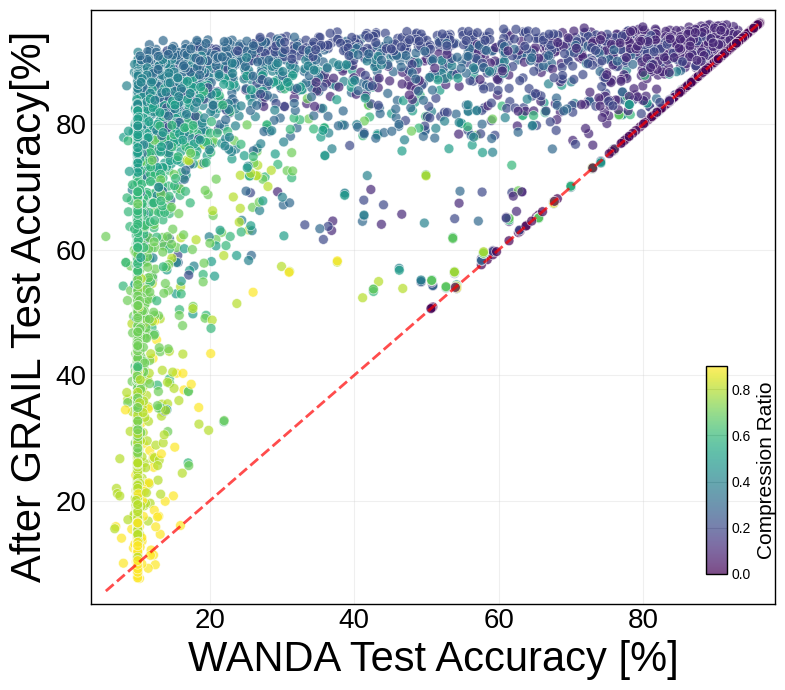}

        \caption*{\textbf{(a) ResNet-18:} \name improves accuracy across pruning strategies.}
    \end{minipage}

    \vspace{1em}

    \begin{minipage}{\textwidth}
        \centering
        \includegraphics[width=0.24\linewidth]{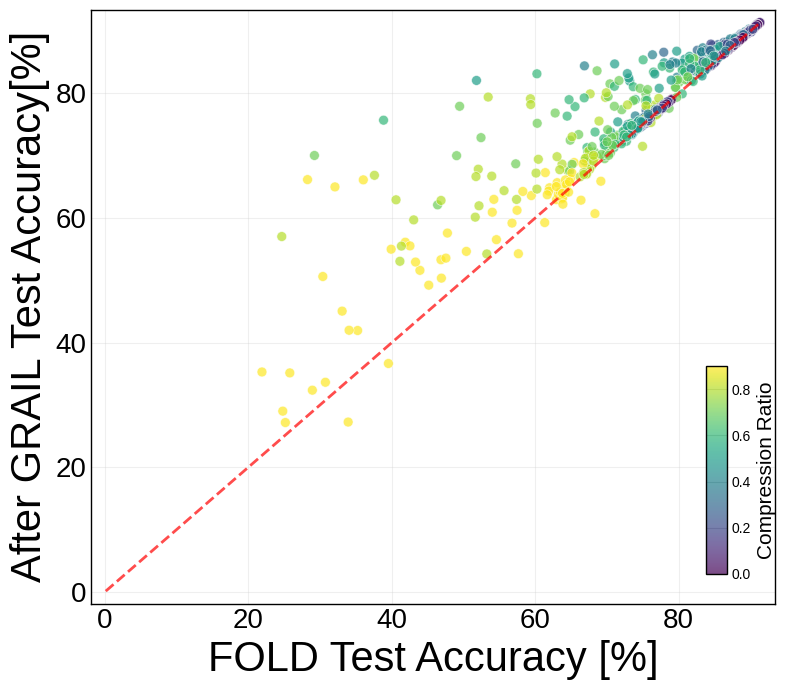}
        \includegraphics[width=0.24\linewidth]{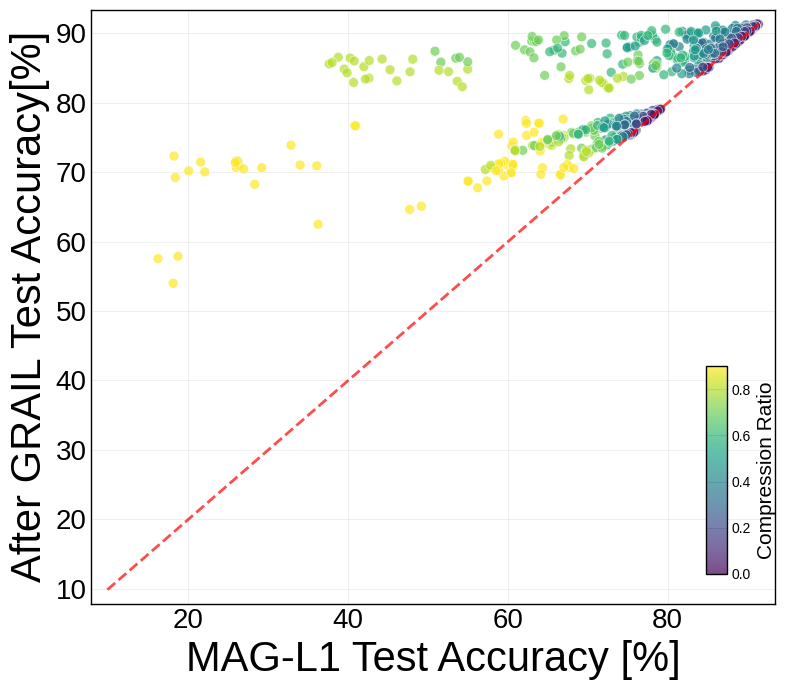}
        \includegraphics[width=0.24\linewidth]{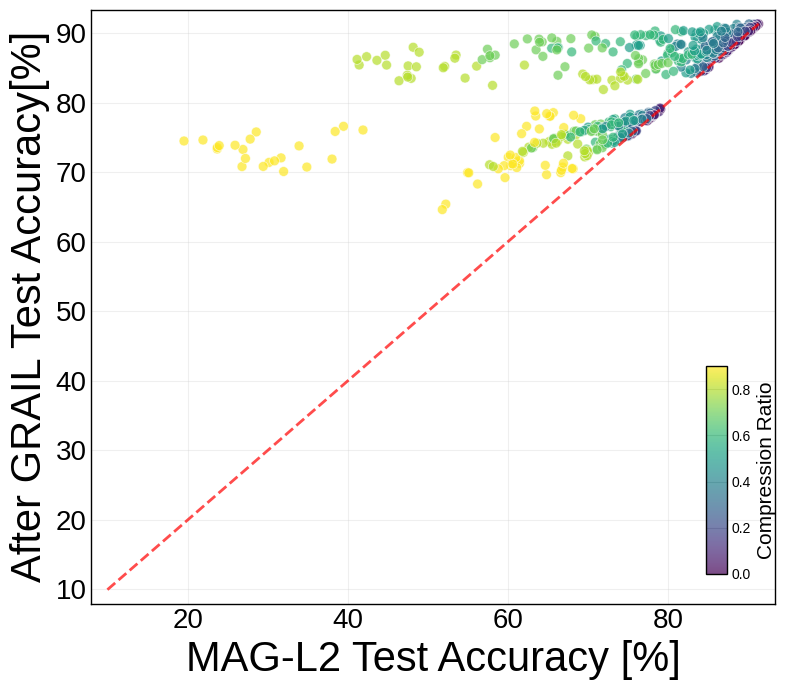}
        \includegraphics[width=0.24\linewidth]{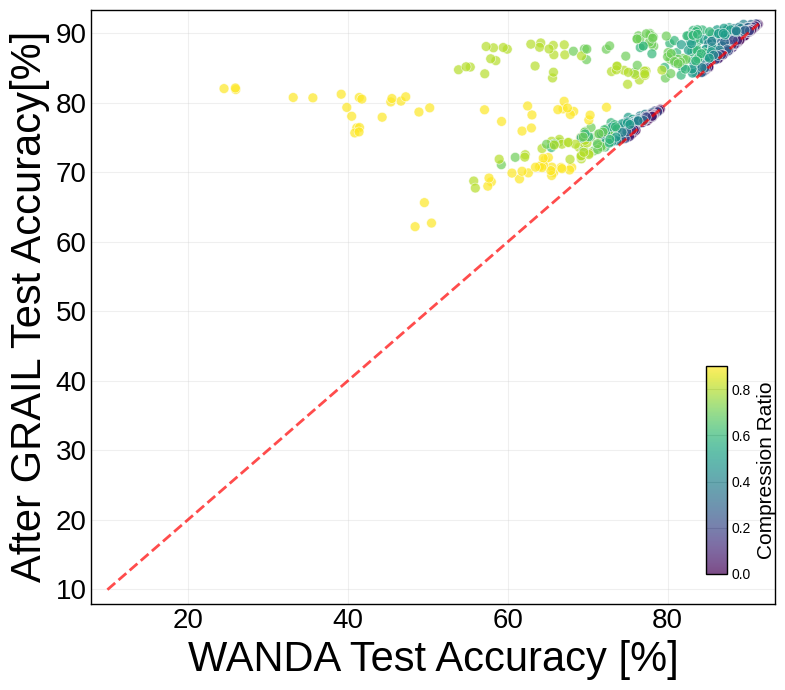}

        \caption*{\textbf{(b) ViT:} Similar upward shift in performance due to \name in transformer architectures.}
    \end{minipage}

    \vspace{1em}

    \begin{minipage}{\textwidth}
        \centering
        \includegraphics[width=0.24\linewidth]{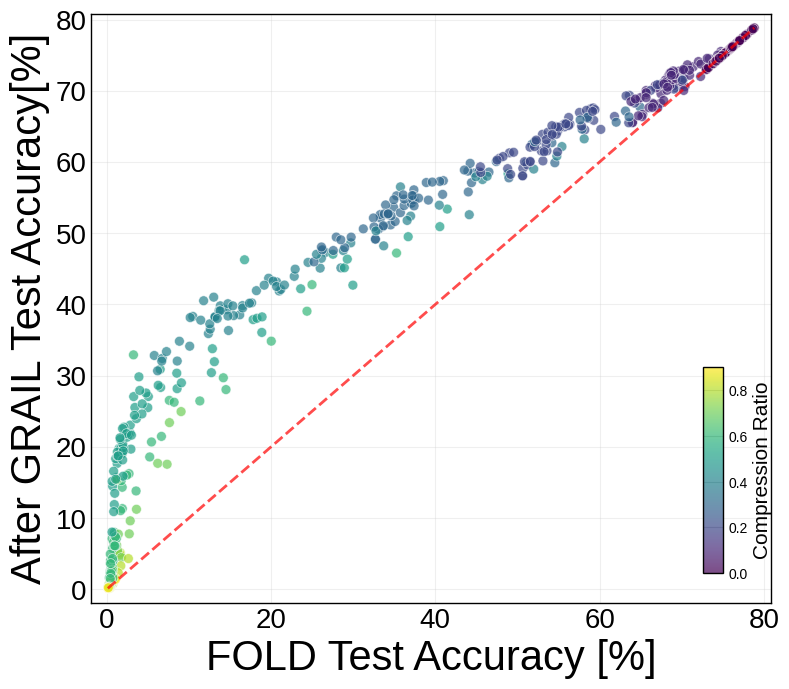}
        \includegraphics[width=0.24\linewidth]{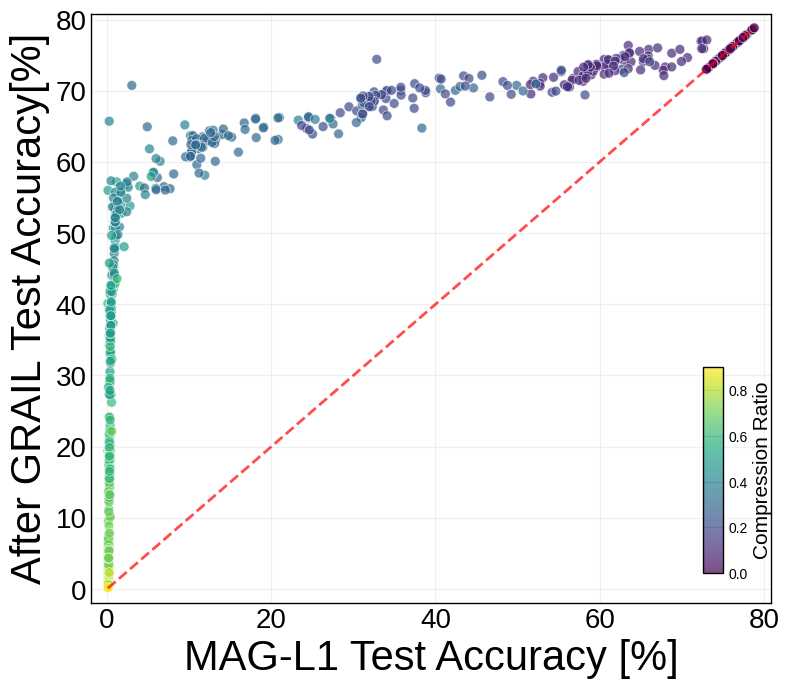}
        \includegraphics[width=0.24\linewidth]{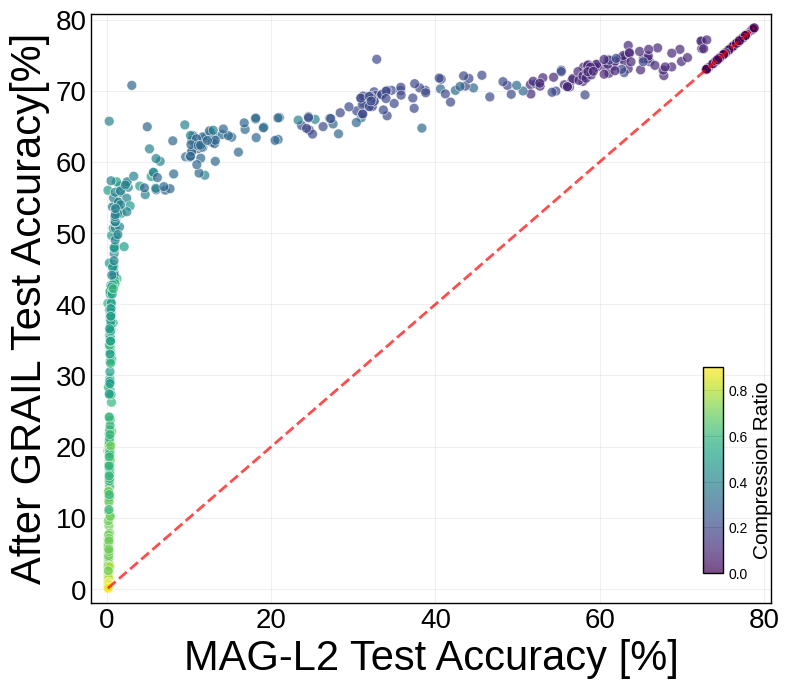}
        \includegraphics[width=0.24\linewidth]{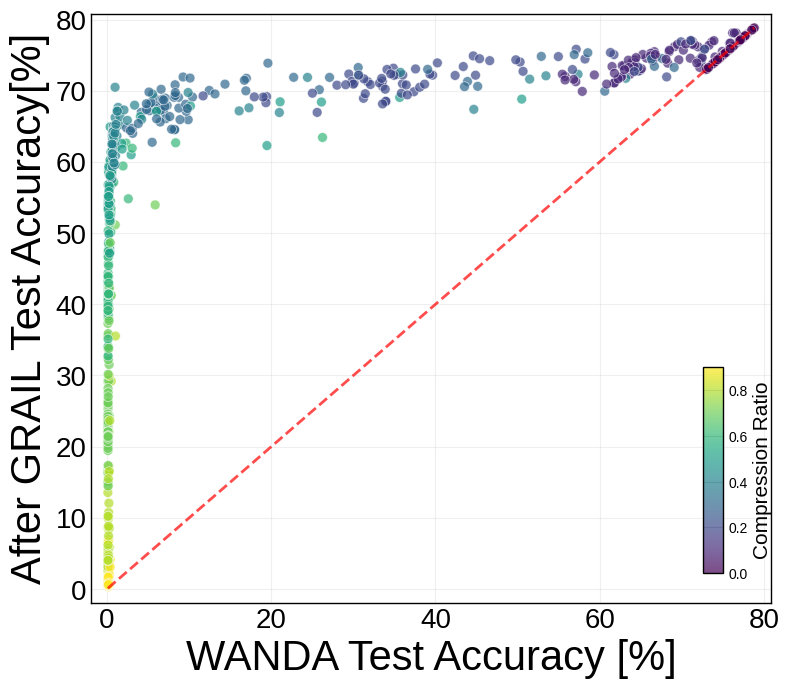}

        \caption*{\textbf{(c) CLIP:} \name consistently boosts encoder representations.}
    \end{minipage}

    \caption{\textbf{\name across pruning and folding compression for ResNet-18, ViT-B/32, and CLIP ViT-B/32.} From left to right: Folding, Mag-L1, Mag-L2, and Wanda. Across all architectures and reduction methods, \name produces a consistent upward shift in accuracy, demonstrating robust post-compression recovery.}

    \label{fig:compensation_results}
\end{figure*}

\end{document}